\documentclass[conf]{new-aiaa}
\usepackage[utf8]{inputenc}
\usepackage{textcomp}
\usepackage{blindtext}
\usepackage{hyperref}
\usepackage[pdftex]{graphicx,xcolor}
\usepackage{pgf}
\usepackage{pgfplots}
\pgfplotsset{compat=newest}
\usepgfplotslibrary{fillbetween}
\usepgfplotslibrary{external} 
\usepackage{tikz} 
\usepackage{tikzscale}
\usetikzlibrary{fadings,plotmarks,shapes,arrows.meta,decorations.markings,patterns,calc,arrows,positioning,fit,backgrounds}
\usepackage{aircraftshapes}
\usepackage{graphicx}
\usepackage{amsmath}
\usepackage[version=4]{mhchem}
\usepackage{siunitx}
\usepackage{longtable,tabularx}
\usepackage[document,bottom]{ragged2e}
\usepackage{subcaption}
\newsavebox{\imagebox}
\usepackage{multicol}
\usepackage{longtable}
\usepackage{bm}
\allowdisplaybreaks
\pgfplotsset{
	colormap={blackwhite}{gray(0cm)=(0); gray(1cm)=(1)}
}
\tikzset{cross/.style={cross out, draw=black, minimum size=2*(#1-\pgflinewidth), inner sep=0pt, outer sep=0pt},
	cross/.default={0.2cm}}
\pgfdeclareradialshading{arr_reward_shade}{\pgfpoint{0cm}{0cm}}%
{
	rgb(0cm)=(0.55,0.8,0.35); 
	rgb(0.5cm)=(0.6,0.8,0.3);
	rgb(1.7cm)=(0.92,0.9 ,0.38); 
	rgb(1.9cm)=(0.99,0.99,0.9);
	rgb(2.1cm)=(1,1,1)
}

\pgfdeclarefading{fading3}{\pgfuseshading{arr_reward_shade}}

\pgfdeclareradialshading{cav_reward_shade}{\pgfpoint{0cm}{0cm}}%
{rgb(0cm)=(1,0.2,0.1); 
	rgb(1.45cm)=(1,0.5,0.2); 
	rgb(2.6cm)=(1,0.9 ,0.5);  
	rgb(3.0cm)=(0.99,0.99,0.93);
	rgb(3.2cm)=(1,1,1)
}

\pgfdeclareradialshading{aircraft_background}{\pgfpoint{0}{0}}%
{color(0)=(pgftransparent!0);
	color(15)=(pgftransparent!100)}

\pgfdeclareradialshading{sphere}{\pgfpoint{0cm}{0cm}}%
{rgb(0cm)=(0.4,0.4,0.4); 
	rgb(0.2cm)=(0.5,0.5,0.5); 
	rgb(0.6cm)=(1,0.9,0.8); 
	rgb(0.80cm)=(1,1,1);
	rgb(1.0cm)=(1,1,1)}

\pgfdeclareradialshading{sphere2}{\pgfpoint{0cm}{0cm}}%
{rgb(0cm)=(0.4,0.4,0.4); 
	rgb(0.2cm)=(0.5,0.5,0.5); 
	rgb(0.6cm)=(0.8,0.8,0.8); 
	rgb(0.85cm)=(1,1,1);
	rgb(1.0cm)=(1,1,1)}  
\tikzstyle{block} = [draw, rectangle, minimum height=3em, minimum width=4em, align=center,thick]
\pgfdeclarelayer{background}
\pgfdeclarelayer{bbackground}
\pgfdeclarelayer{bbbackground}
\pgfdeclarelayer{foreground}
\pgfsetlayers{bbbackground,bbackground,background,main,foreground}

\pgfplotsset{
	/pgfplots/colormap={custom1}{[1cm]rgb255(0.0cm)=(82,27,53) rgb255(1cm)=(163,63,49) rgb255(2cm)=(204,76,41) rgb255(3cm)=(247,188,37) rgb255(4cm)=(241,240,155)}
}

\pgfplotsset{
	/pgfplots/colormap={custom2}{[1cm]rgb255(0cm)=(0,44,103) rgb255(1cm)=(8,69,148) rgb255(2cm)=(33,113,181) rgb255(3cm)=(66,146,198) rgb255(4cm)=(107,174,214) rgb255(5cm)=(158,202,225) rgb255(6cm)=(198,219,239) rgb255(7cm)=(222,235,247) rgb255(8cm)=(247,251,255) rgb255(9cm)=(255,255,255)}
}

\pgfplotsset{
	/pgfplots/colormap={custom2b}{[1cm]rgb255(0cm)=(8,69,148) rgb255(1cm)=(33,113,181) rgb255(2cm)=(66,146,198) rgb255(3cm)=(107,174,214) rgb255(4cm)=(158,202,225) rgb255(5cm)=(198,219,239) rgb255(6cm)=(222,235,247) rgb255(7cm)=(247,251,255) rgb255(8cm)=(255,255,255)}
}

\pgfplotsset{
	/pgfplots/colormap={custom2c}{[1cm]rgb255(0cm)=(11,28,64) rgb255(6cm)=(81,118,181) rgb255(14cm)=(225,231,242) rgb255(16cm)=(235,239,247)}
}

\pgfplotsset{
	/pgfplots/colormap={custom3}{[1cm]rgb255(0.0cm)=(8,8,158) rgb255(1cm)=(43,140,190) rgb255(2cm)=(78,179,211) rgb255(3cm)=(123,204,196) rgb255(4cm)=(168,221,181) rgb255(5cm)=(204,235,197) rgb255(6cm)=(224,243,219) rgb255(7cm)=(247,252,240)}
}

\pgfplotsset{
	/pgfplots/colormap={custom4}{[1cm]rgb255(0.0cm)=(19,27,74) rgb255(1cm)=(12,44,132) rgb255(2cm)=(34,94,168) rgb255(3cm)=(29,145,192) rgb255(4cm)=(65,182,196) rgb255(5cm)=(127,205,187) rgb255(6cm)=(199,233,180) rgb255(7cm)=(237,248,177) rgb255(8cm)=(255,255,217)}
}

\pgfplotsset{
	/pgfplots/colormap={custom4b}{[1cm]rgb255(0cm)=(12,44,132) rgb255(1cm)=(34,94,168) rgb255(2cm)=(29,145,192) rgb255(3cm)=(65,182,196) rgb255(4cm)=(127,205,187) rgb255(5cm)=(199,233,180) rgb255(6cm)=(237,248,177) rgb255(7cm)=(255,255,217)}
}

\pgfplotsset{
	/pgfplots/colormap={custom5}{[1cm]rgb255(0.0cm)=(122,0,55) rgb255(1cm)=(177,0,38) rgb255(2cm)=(227,26,28) rgb255(3cm)=(252,78,42) rgb255(4cm)=(253,141,60) rgb255(5cm)=(254,178,76) rgb255(6cm)=(254,217,118) rgb255(7cm)=(255,237,160) rgb255(8cm)=(255,255,204)}
}

\pgfplotsset{
	/pgfplots/colormap={hot2}{[1cm]rgb255(0cm)=(0,0,0) rgb255(3cm)=(255,0,0)
		rgb255(6cm)=(255,255,0) rgb255(8cm)=(255,255,255)}
}

\title{Long Short-Term Memory for Spatial Encoding in Multi-Agent Path Planning\footnotetext{\textit{For the source code, please see {\footnotesize\url{https://github.com/MarcSchlichting/LSTMSpatialEncoding}}}}}

\author{Marc R. Schlichting \footnote{Undergraduate Student, Institute of Flight Mechanics and Controls, now Graduate Student at Stanford University, AIAA Student Member.}, Stefan Notter\footnote{Research Associate, Institute of Flight Mechanics and Controls.}, and Walter Fichter\footnote{Professor, Institute of Flight Mechanics and Controls, AIAA Associate Fellow.}}
\affil{University of Stuttgart, Stuttgart, Germany, 70569}

\newcommand*\summation{\tikz[baseline=(char.base)]{
		\node[shape=circle,draw,inner sep=1pt] (char) {\scriptsize$+$};}}

\newcommand*\hadamard{\tikz[baseline=(char.base)]{
		\node[shape=circle,draw,inner sep=1pt] (char) {$\circ$};}}
\begin{document}
	
	\maketitle
	
	\begin{abstract}
		Reinforcement learning-based path planning for multi-agent systems of varying size constitutes a research topic with increasing significance as progress in domains such as urban air mobility and autonomous aerial vehicles continues. Reinforcement learning with continuous state and action spaces is used to train a policy network that accommodates desirable path planning behaviors and can be used for time-critical applications. A Long Short-Term Memory module is proposed to encode an unspecified number of states for a varying, indefinite number of agents. The described training strategies and policy architecture lead to a guidance that scales to an infinite number of agents and unlimited physical dimensions, although training takes place at a smaller scale. The guidance is implemented on a low-cost, off-the-shelf onboard computer. The feasibility of the proposed approach is validated by presenting flight test results of up to four drones, autonomously navigating collision-free in a real-world environment.
	\end{abstract}
	
	\section*{Nomenclature}
\begin{multicols}{2}
	{\renewcommand\arraystretch{1.0}
		\noindent\begin{tabular}{@{}l @{\quad=\quad} l@{}}
		$\mathcal{A}$ & Action Space\\
		$a$ & Action \\
		$\bm{c}$ & Cell Values \\
		$\mathfrak{d}$ & Euclidean Distance $\left[\text{m}\right]$\\
		$\delta$ & Radius of Delta Zone $\left[\text{m}\right]$\\
		$\varepsilon$ & Radius of Epsilon Zone $\left[\text{m}\right]$\\
		$\mathbb{E}$ & Estimated Value\\
		$\bm{f}$ & Forget Gate\\
		$\bm{h}$ & Hidden State \\
		$\bm{i}$ & Input Gate \\
		$i$ & Index of Own Agent\\
		$j$ & Index of Other Agents \\
		$\bm{o}$ & Output Gate \\	
		\end{tabular}}
	
	{\renewcommand\arraystretch{1.0}
		\noindent\begin{tabular}{@{}l @{\quad=\quad} l@{}}		
		$\pi$ & Policy \\
		$R$ & Reward \\
		$\mathcal{R}$ & Maximum Reward \\
		$\bm{r}$ & Position $\left[\text{m}\right]$\\
		$\mathcal{S}$ & State Space \\
		$s$ & State \\
		$\sigma$ & Sigmoid Function\\
		$\bm{U} / \bm{W}$ & Weight Matrices \\
		$\mathcal{V}$ & Set of Vehicles\\
		$\bm{x}$ & Input Tuple \\
		$\psi$ & Bearing $\left[\text{rad}\right]$\\
		$\Omega$ & Physical Space \\
		$\circ$ & Hadamard Product 
\end{tabular}}
\end{multicols}

	\section{Introduction}\label{chap:introduction}
	\lettrine{P}{ath} planning is crucial for many applications in the field of robotics but receives a special interest for large, autonomous multi-agent systems operating under dynamically changing, uncertain situations~\cite{Katz.2019}. Given such conditions, path planning needs to ensure the fulfillment of the mission goals while dealing with other participants and the environment at the same time. Since the two main mission goals for path planning---trajectory tracking and collision avoidance---have different objectives and approaches for solving them \cite{Kuchar.2000}, balancing the two objectives is a key aspect for every multi-agent path planning algorithm. By assuming a unified strategy and goal for all participants, the path planning problem becomes a cooperative task~\cite{Beard.2006,Kingston.2008}. Solving a problem using a holistic perspective can sacrifice the performance of individual agents to improve the system's overall performance. In biology, this paradigm is known as swarming behavior. 
	
	Optimal path planning is dominated by search or optimization algorithms such as \textit{Rapidly Exploring Random Trees}~(RRTs)~\cite{LaValle.1998}, \textit{Probabilistic Road Maps}~(PRMs)~\cite{Kavraki.1996}, and \textit{Model Predictive Control}~(MPC)~\cite{Mayne.2000,Joos.2011,Ji.2017}. However, these approaches exhibit significant drawbacks when applied to complex and dense physical spaces and an increased number of agents, as the required computation times can render these methods unsuitable for real-time applications such as decentralized control a swarm of unmanned aerial vehicles~(UAVs). 
	
	MPC aims for minimizing an objective function over a receding horizon. The efficiency of a control strategy based on real-time optimization is highly dependent on---and often limited by---the optimization algorithm used. To solve the time-variant optimization problem for path planning with a varying number of agents, solvers that can cope with partially chaotic systems are required. Mixed-integer linear programming is a method that is often used in this context and facilitates the implementation of restricted flight zones, as well as the incorporation of linear, dynamic constraints~\cite{Bellingham.2002,Richards.2002}. However, due to the online optimization, MPC requires computational power that often is beyond the capabilities of onboard computers, especially if the model used for optimization is sophisticated. 
	
	RRTs also belong to the group of online optimization algorithms. Different variants that extend the originally proposed algorithm by optimality exist, but they all are non-heuristic and, therefore, their use for time-variant real-time applications where the difficulty of the optimization problem changes with every time step can be problematic \cite{Karaman.2011}. An intermediate approach between online and offline methods is represented by PRMs~\cite{Kavraki.1996}, where the graph containing all possible states and their transitions is calculated offline while planning on the pre-computed graph takes place online. The most extreme form of offline methods can be found in state-action score tables with pre-computed optimal solutions for given situations~\cite{Kochenderfer.2012}. A major drawback of table-based methods, however, is the amount of memory required to store the results~\cite{Julian.2018a}. 
	
	Path planning algorithms can be divided into two categories. Algorithms affiliated with the first category generate a complete trajectory as an output~\cite{Watanabe.2018,Inoue.2019}. This trajectory can be continuously updated if necessary; however, only the first step is used for actuator commands. The second category does not generate a trajectory directly; it only issues the command for the current time step~\cite{Julian.2018a,Notter.2019}. A complete trajectory can be obtained by the propagation of the system. Artificial potential fields, for which the agent's actions are based on the gradient of the potential at the current position represent an example of this second category~\cite{Yang.2000,Khatib.1986}. The second approach is advantageous since it avoids the generation of unnecessary data and is easier to implement.
	
	Since all previously described methods are subject to either computational or memory limitations, we look at deep learning using \textit{artificial neural networks}~(ANNs). ANNs represent a class of highly adaptable functions that can be tailored to application-specific requirements. Applications include numerous tasks in the field of robotics such as the replacement of conventional controllers~\cite{Aamir.2013} or complex tasks such as lane-keeping in autonomous driving based on raw sensory input~\cite{Pomerlau.1989}. Using ANNs as a non-linear approximation for a path planning design model has not only been proven functional~\cite{Richards.2010,Watanabe.2018}, but also provides a heuristic approach for near-optimal path planning. Optimization with regard to a mission objective takes place while training the ANN. At the evaluation, an estimate of the optimal behavior is obtained; however, hard time constraints can be observed~\cite{Qu.2009}. 
	
	Different approaches exist to train the parameters of an ANN-based path planning model, which is commonly referred to as a policy. Supervised learning-based methods are used to mimic the behavior of an expert. This expert can be a human pilot~\cite{Abbeel.2010,Tran.2015} or a numerically determined optimal solution~\cite{Julian.2017,Watanabe.2018}. RRTs can also be used as optimal solutions~\cite{Inoue.2019}. A hybrid strategy wherein an ANN works alongside a conventional controller is reported to be successfully employed by~\cite{Matsuura.2007}. Since all approaches using supervised learning are reliant on extensive datasets and the imitation of the recorded behavior is generally less powerful than the method on which it is based, we explore reinforcement learning, an approach where an optimal behavior is obtained purely by interacting with the environment~\cite{Sutton.2018,Mnih.2015}. 
	
	Reinforcement learning can successfully be applied to path planning for multi-agent systems~\cite{Julian.2018a,Julian.2019,Beard.2006,Kingston.2008}. It has been used for coordinated wildfire surveillance~\cite{Julian.2019} as well as for collision avoidance~\cite{Julian.2018a}. Both scenarios use multiple agents; however, the number of agents has to be constant to comply with the nature of most ANN architectures that expect a time-invariant state space. Dealing with a varying amount of agents requires the implementation of an encoding mechanism or a decomposition into tasks with a constant number of agents and recombination into the multi-agent scenario~\cite{Ong.2015}. Most often, occupation maps are used for this purpose~\cite{Watanabe.2018,Inoue.2019,Munos.2002}. Using discrete occupation maps for spatial encoding, however, not only limits the performance of a downstream algorithm for a continuous path-planning problem but is also inefficient with regard to memory usage for a sufficient resolution of a large state space.
	
	Following the approach of relative positions of the agents~\cite{Julian.2018a,Julian.2019} yields a higher information density compared to occupation maps. However, dealing with a varying number of agents in that context has not been investigated so far. Such an approach would make maneuvering in cluttered situations more reliable since more relevant information can be passed efficiently to the policy. Overall, the decision-making process becomes more informed. Nevertheless, an encoding mechanism is required to process the variably-sized input. 
	
	The contribution of this paper is a multi-agent path planning approach that is memory-efficient and scalable concerning the number of agents. This allows to overcome limitations that are given under practical implementation conditions. It is proposed to apply a \textit{Long Short-Term Memory} module~(LSTM)~\cite{Hochreiter.1997} for encoding the relative vehicle positions, followed by a non-recurrent ANN to transform the encoded states alongside other relevant information into actions\footnote{A similar approach was developed independently and was presented at the same time on the AIAA SciTech Forum 2021 \cite{Brittain2021}.}. This novel approach to spatial encoding is both memory-efficient and scalable concerning the number of agents. Therefore, employing a ``Long Short-Term Memory for Spatial Encoding'' combines the advantages of classical encoding mechanisms while not suffering from the respective drawbacks. Concerning the application of multi-agent path planning, the overall control policy makes the vehicles reach their target positions in minimum time while avoiding collisions and excessive accelerations. The authors demonstrate the scalability of the approach to a nearly unlimited number of agents with minimal training required. Finally, the algorithm is implemented on low-cost, off-the-shelf embedded hardware for quadrotor drones. Flight test results of up to four drones navigating autonomously and without collision demonstrate the feasibility and performance of the proposed approach in a real-world environment.
	
	\section{Deep Reinforcement Learning for Path Planning}
	
	\subsection{Assumptions and Equations of Motion}
	A flat, non-rotating Earth is assumed. All vehicles are treated as point masses; thus the equations of motion simplify to:
	\begin{equation}
		\bm{\dot{r}}=\bm{v}
		\label{eq:eq_mo}
	\end{equation}
	\par The model is simulated using Euler's method for fast evaluation. All observations are limited to the horizontal plane of motion. Using a simplified model for the purpose of training accelerates the training process significantly and still results in the desired behavior when interacting with the original environment, provided that reasonable assumptions have been made \cite{Kuwata.2004}.
	
	\subsection{Reinforcement Learning: Actions, Observations, and Policy}
\begin{figure}[t!]
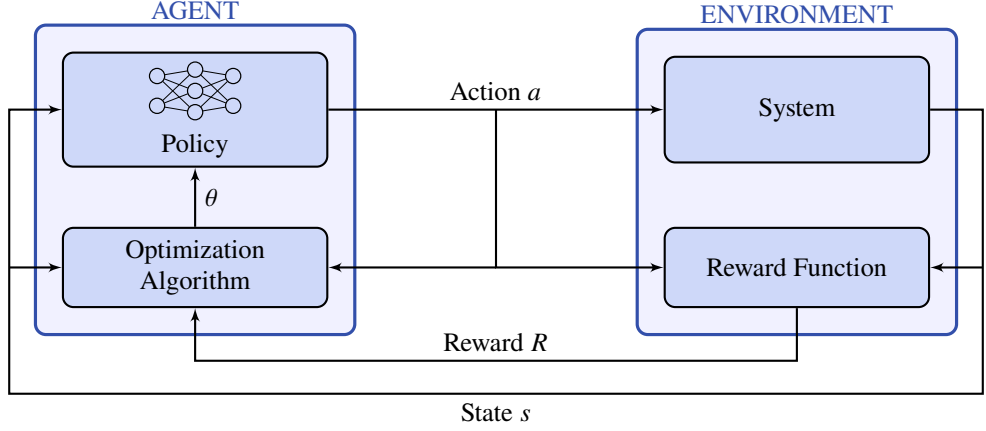

		\vskip -5.6mm
		\centering
		\include{extended_agent_env}
		\caption{Interaction between the agent and the environment in Reinforcement learning.}
		\label{fig:agent_environment}
	\end{figure}

	Reinforcement learning is a biologically inspired method with a characteristic interaction between an agent and the environment. The agent submits actions $a$ to the environment, and a reward $R$---a quantity describing the compliance with a previously agreed-upon mission objective---is returned to the agent together with the new states. An optimization algorithm alters the parameters $\theta$ of the agent's policy in an attempt to increase the reward. The interaction between an agent and the environment is illustrated in Fig. \ref{fig:agent_environment}.
	\par For the path-planning application discussed, all actions are expressed as commanded velocities in a body-fixed coordinate system, which is advantageous because it reduces the complexity of the simulation model. However, this procedure does not resemble reality, where forces and torques are controlled rather than velocities. Nevertheless, as the policy is expected to behave in an acceleration-minimizing manner due to the reward function shaped as described in section \ref{chap:reward_function}, the approach seems reasonable. The set of all possible actions is denoted as $\mathcal{A}$, where $a\in\mathcal{A}$ refers to a specific action.
	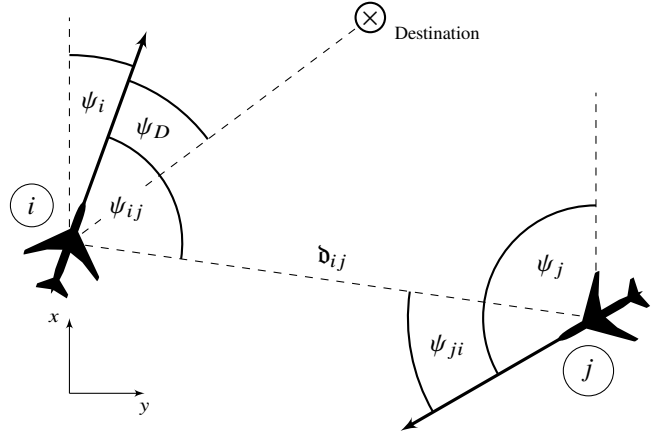
\begin{figure}[t!]
		\centering
\begin{tikzpicture}
\node [aircraft top,fill=black,minimum width=1.3cm,rotate=70] at (0,0) (i) {};
\node[draw,circle] at (-0.5,0.5) (i_label){$i$};
\node [aircraft top,fill=black,minimum width=1.3cm,rotate=210] at (7,-1) (j) {};
\node[draw,circle] at (6.9,-1.7) (j_label){$j$};
\draw[thick] ([shift=(-8.13:1.5cm)]i) arc (-8.13:70:1.5cm);
\draw[thick] ([shift=(70:2.5cm)]i) arc (70:90:2.5cm);
\draw[draw,dashed] (i) -- ($(i)+(90:3)$);
\draw[thick] ([shift=(90:1.5cm)]j) arc (90:210:1.5cm);
\draw[draw,dashed] (j) -- ($(j)+(90:3)$);
\draw[thick] ([shift=(171.87:2.5cm)]j) arc (171.87:210:2.5cm);
\draw[draw,-latex',very thick] (i) -- ($(i)+(70:3)$);
\draw[draw,-latex',very thick] (j) -- ($(j)+(-150:3)$);
\draw[draw,dashed] (i) -- node[above, pos=0.5]{$\mathfrak{d}_{ij}$}(j);

\node[] at ($(i)+(80:1.9cm)$) (psi_i){$\psi_{i}$};
\node[] at ($(j)+(90+40.935:0.9cm)$) (psi_j){$\psi_{j}$};
\node[] at ($(j)+(190.935:2.0cm)$) (psi_ji){$\psi_{ji}$};
\draw[draw,-latex'] (0,-2) -- node[pos=1,below]{\scriptsize $y$}(1,-2);
\draw[draw,-latex'] (0,-2) -- node[pos=1,left]{\scriptsize $x$}(0,-1);

\node [draw,thick,inner sep=1pt,circle] at (4,3) (target) {\textbf{$\times$}};
\node [anchor = west] at ($(target)-(-0.2,0.2)$) (label-target) {\scriptsize Destination};
\draw[draw,dashed] (i) -- (target);

\node[fill=white] at ($(i)+(30.935:0.9cm)$) (psi_ij){$\psi_{ij}$};

\draw[thick] ([shift=(36.8699:2.3cm)]i) arc (36.8699:70:2.3cm);

\node[] at ($(i)+(53.4349:1.9cm)$) (psi_d){$\psi_{D}$};

\end{tikzpicture}
		\caption{Describing quantities: Relative positions between two agents. Positive angles are clockwise.}
		\label{fig:relative_angles}
	\end{figure}
	\par Observations regarding each other agent $j$ from the perspective of agent $i$ are described using a tuple of three quantities: $\mathfrak{d}_{ij}$ referring to the distance between the two agents, $\psi_{ij}$ denoting the angular position of $j$ as seen from $i$ in the direction of movement, and $\psi_{ji}$ describing the included angle between the velocity vector of $j$ and the line of sight between $i$ and $j$. All described quantities are depicted in Fig. \ref{fig:relative_angles}. The set of all possible states, the state space, is referred to as $\mathcal{S}$, where $s\in\mathcal{S}$ denotes a specific state. The policy itself takes observations\footnote{We assume observations equaling states, i.e., a fully observable decision process, herein.} as input and maps them to the actions as output: $a=f(s)$.

	\subsection{Proximal Policy Optimization as the Training Algorithm}
	The goal of reinforcement learning is to train an agent's policy $\pi$ in such a way that actions that contribute to higher rewards will be more likely in future evaluations of the policy than those that lead to a lower reward. It is the agent's objective to maximize the reward. In order to obtain a policy that not only delivers desirable results for the immediate reward, but also takes future evaluations into account, the concept of a discounted reward $\eta$ is defined as:
	\begin{equation}
		\eta\left(s_0,a_0,s_1,\ldots\right)=\sum\limits_{t=0}^{\infty}\gamma^tR\left(s_t\right)\label{eq:discounted_reward}
	\end{equation}
	where $\gamma\in\left(0,1\right)$ is known as the discount factor. The value of a state $V_\pi$ describes the expected discounted reward when applying $\pi$ at that particular state, taking transition probabilities into account. Additionally, the value of an action $Q_\pi$ measures the expected discounted reward for a given action and a given state \cite{Sutton.2018}:
	\begin{align}
		V_\pi\left(s_t\right)&=\mathbb{E}_{a_t,s_{t+1},\ldots\sim\pi}\left[\sum\limits_{k=0}^{\infty}\gamma^kR\left(s_{t+k}\right)\right]\label{eq:V_eq}\\
		Q_\pi\left(s_t,a_t\right)&=\mathbb{E}_{s_{t+1},a_{t+1},\ldots\sim\pi}\left[\sum\limits_{k=0}^{\infty}\gamma^kR\left(s_{t+k}\right)\right]\label{eq:Q_eq}
	\end{align}
	The so-called advantage $A$ is the difference between $Q$ and $V$, $A=Q-V$, where $V$ and $Q$ are defined in Eqs. (\ref{eq:V_eq}) and (\ref{eq:Q_eq}), respectively. If $A$ is positive, the action taken is expected to lead to a larger reward than the expected reward for that particular state following policy $\pi$. The opposite applies for a negative advantage.
	\par \textit{Proximal Policy Optimization} (PPO) is a policy gradient-based method. The loss for a gradient descent is defined as:
	\begin{equation}
		L^{PG}\left(\theta\right)=\hat{\mathbb{E}}_t\left[\log\pi_{\theta}\left(a_t|s_t\right)\hat{A}_t\right]\footnote{$\theta$ denotes the parameters (i.e., weights and biases) of a policy. Accordingly, $\pi_\theta$ denotes the policy parametrized by $\theta$. However, it is common to use $\theta$ as an abbreviation for $\pi_{\theta}$.}
		\label{eq_lpg}
	\end{equation}
	\par Given a captured episode of the agent interacting with the environment, the only variable that needs to be estimated in the advantage is the value of the state. $Q$ can be determined using the captured data. A stochastic gradient descent would determine the gradient of the loss with respect to $\theta$ and modify the parameters accordingly. However, the first order derivative can become invalid, especially when the objective features a large curvature. This can lead to unstable behavior given a specific update step. The predecessor of PPO, \textit{Trust Region Policy Optimization} (TRPO), refers to the constrained optimization problem \cite{schulman2015trust}:
	\begin{equation}
		\label{eq:trpo_obj}
		\begin{alignedat}{2}
			&\underset{\theta}{\text{maximize}}& &\quad\hat{\mathbb{E}}_t\left[r_t\left(\theta\right)\cdot\hat{A}_t\right]\\
			&\text{subject to}& &\quad\hat{\mathbb{E}}_t\left[D_{KL}\left[\pi_{\theta_{t-1}}\left(\cdot|s_t\right),\pi_{\theta_{t}}\left(\cdot|s_t\right)\right]\right]\le d
		\end{alignedat} 
	\end{equation}
	where $r_t\left(\theta\right)=\frac{\pi_{\theta_t}\left(a_t|s_t\right)}{\pi_{\theta_{t-1}}\left(a_t|s_t\right)}$, $\hat{A}$ is the estimated advantage, and $D_{KL}$ is the Kullback-Leibler divergence. Hence, TRPO restricts the difference between two policies using the Kullback-Leibler divergence to an elliptical search region. TRPO is based on the idea of a line search along the line connecting the original policy and the policy after an update using natural gradient ascent (restricted by the trust region). Optimizing the surrogate objective stated in Eq. (\ref{eq:trpo_obj}) is equivalent to the more intuitive line search method \cite{schulman2015trust}. However, due to the difficulties in estimating $D_{KL}$ and the detrimental effects of extremely positive policy update estimates through the ratio $r\left(\theta\right)$, the constrained problem was replaced by an unconstrained optimization problem in PPO where $r_t\left(\theta\right)$ is clipped instead and a surrogate objective is defined as \cite{schulman2017proximal}:
	\begin{equation}		
		L^{CLIP}\left(\theta\right)=\hat{\mathbb{E}}_t\left[\min\left(r_t\left(\theta\right)\hat{A}_t,\text{clip}\left(r_t\left(\theta\right),1-\epsilon,1+\epsilon\right)\hat{A}_t\right)\right]\label{eq:l_clip}
	\end{equation}
	where $\epsilon$ is a hyperparameter and $\mathrm{clip}(x,a,b)$ is the clipping function where $\mathrm{clip}(x,a,b)=a$ if $x<a$, $\mathrm{clip}(x,a,b)=b$ if $x>b$, and $\mathrm{clip}(x,a,b)=x$ otherwise. For the training of the policy's parameters, we use the Adam optimizer. With the advantage of the loss function being estimated based on a batch of sampled tuples of state, action, and reward received, the estimated loss becomes a random variable itself. We employ Adam for optimizing the parameters on the control policy. Adam is a well-established algorithm for optimization that uses momentum and efficient step size annealing to improve the convergence of stochastic gradient descent \cite{Kingma.2014}.
	\subsection{Reward Function}
	\label{chap:reward_function}
	The reward function is used to outline a desirable agent behavior and plays an essential role for the training (Eqs. (\ref{eq:discounted_reward}) to (\ref{eq:Q_eq})) and the behavior of the agent. A negative reward $R\in\mathbb{R}^-$ is associated with an undesirable state or action, whereas a positive reward $R\in\mathbb{R}^+$ indicates a preferable action or state. The reward function is formed as the weighted sum of four behavioral patterns:
	\begin{enumerate}
		\item \textbf{Arrival at target position:}
		\par For the arrival reward, an $\varepsilon$-zone around the target position $\bm{r}_D$ of radius $\varepsilon_{ARR}$ is defined:
		\begin{equation}
			\Omega_{\varepsilon,ARR}:=\left(\forall\bm{r}:\left(\mathfrak{d}_D\le\varepsilon_{ARR}\right)=\text{True}\right)
		\end{equation}
		where $\mathfrak{d}_D=\left\Vert\bm{r}_D-\bm{r}\right\Vert$ is the Euclidean distance between the agent's current position and the destination. Let $\mathcal{R}_{ARR}\in\mathbb{R}^+$ denote a constant reward that is rewarded once the agent is inside its respective $\Omega_{\varepsilon,ARR}$, then the arrival reward $R_{ARR}$ holds:
		\begin{equation}
			R_{ARR}:=\begin{cases}
				\mathcal{R}_{ARR}&\forall\bm{r}\in\Omega_{\varepsilon,ARR}\\
				0&\forall\bm{r}\not\in\Omega_{\varepsilon,ARR}
			\end{cases}\label{eq:arr_reward}
		\end{equation}
		An example portraying different arrival reward outcomes can be seen in Fig. \ref{fig:arr_reward}. $i_1$ is located inside the $\varepsilon$-region, therefore receiving the full arrival reward. $i_2$ and $i_3$ are both located outside the $\varepsilon$-region and consequently receive an arrival reward of zero, despite $i_2$ being closer to the $\varepsilon$-region than $i_3$. The binary nature of the arrival reward makes the training more stable.
		\item \textbf{Collision avoidance:}
		\par In a similar manner, $\Omega_{\varepsilon,CAV,j}$ for collision avoidance between vehicle $i$ and $j$ is defined as:
		\begin{equation}
			\Omega_{\varepsilon,CAV,j}:=\left(\forall\bm{r}_i:\left(\mathfrak{d}_{ij}\le\varepsilon_{CAV}\right)=\text{True}\right)
		\end{equation}
		with $\mathfrak{d}_{ij}$ as Euclidean distance between $i$ and $j$. Another region, the so-called $\delta$-region $\Omega_{\delta,CAV,j}$ is defined by:
		\begin{equation}
			\Omega_{\delta,CAV,j}:=\left(\forall\bm{r}_i:\left(\left(\mathfrak{d}_{ij}>\varepsilon_{CAV}\right)\land\left(\mathfrak{d}_{ij}\le\delta_{CAV}\right)\right)=\text{True}\right)
		\end{equation}
		where $\delta_{CAV}>\varepsilon_{CAV}$. Entering $\Omega_{\varepsilon,CAV,j}$ is prohibited, while entering $\Omega_{\delta,CAV,j}$ is not preferred but tolerated. The collision reward for the vehicle pair $i$ and $j$ then reads:
		\begin{equation}
			R_{CAV,i,j}:=\begin{cases}
				\mathcal{R}_{CAV}&\forall\bm{r}_i\in\Omega_{\varepsilon,CAV,j}\\
				\mathcal{R}_{CAV}\cdot f\left(\mathfrak{d}_{ij}\right)&\forall\bm{r}_i\in\Omega_{\delta,CAV,j}\\
				0&\forall\bm{r}_i\not\in(\Omega_{\varepsilon,CAV,j}\cup\Omega_{\delta,CAV,j})
			\end{cases}\label{eq:cav_reward_cont}
		\end{equation}
		where $\mathcal{R}_{CAV}\in\mathbb{R}^-$ defines the maximum negative collision avoidance reward. $f$ is an arbitrary function that is used as the transition between the boundaries $\partial\Omega_{\varepsilon,CAV}$ and $\partial_o\Omega_{\delta,CAV}$, in the most simple case, a linear transition. The final collision reward for vehicle $i$ then reads:
		\begin{equation}
			R_{CAV,i}=\sum_{j}R_{CAV,i,j}\quad \text{with}\quad(j\in\mathcal{V})\land (i\ne j)
		\end{equation}
		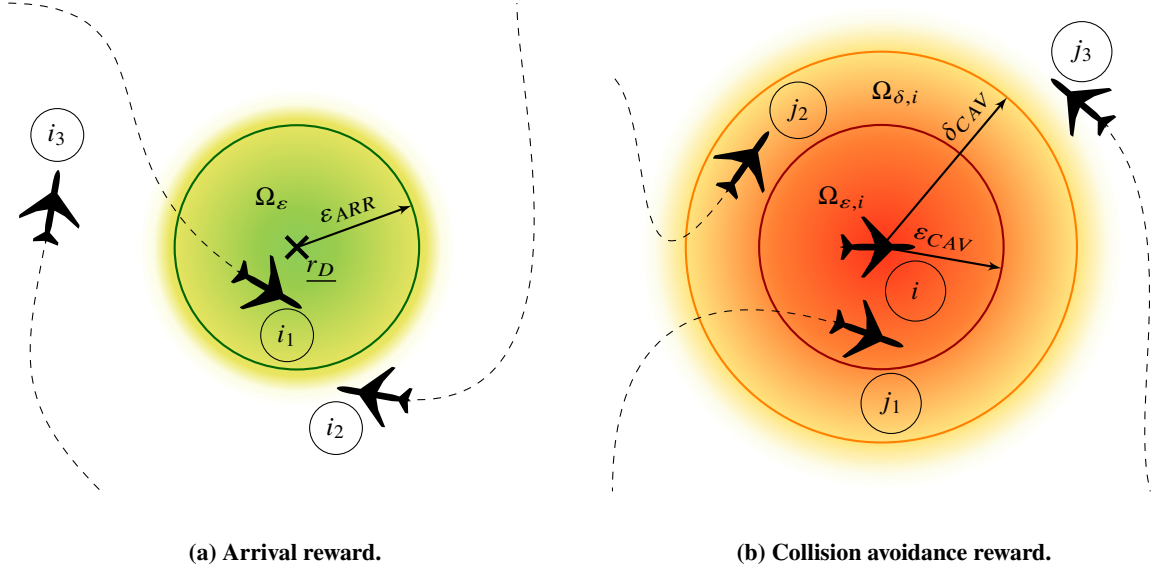
\begin{figure}[t!]
			\centering
			\begin{subfigure}[c]{0.455\textwidth}
				\centering
				\begin{tikzpicture}[scale=0.65]
\pgfuseshading{arr_reward_shade}

\fill[opacity=0] (0cm,0cm) circle (5cm);
\draw[green!40!black,thick] (0cm,0cm) circle (2.5cm);
\draw (0cm,0cm) node[cross,ultra thick] {};
\node[] at (0.5cm,-0.5cm) (label_destination) {$\underline{r_D}$};

\draw[thick,-latex'] (0cm,0cm) -- node[pos=0.5,above,rotate=20]{$\varepsilon_{ARR}$}($(0,0)+(20:2.5)$);

\node [aircraft top,fill=black,minimum width=1cm,rotate=-30] at (-0.5,-0.9) (ac1) {};
\node[circle,draw] at ($(ac1)+(0.3,-0.9)$) {$i_1$};
\node [aircraft top,fill=black,minimum width=1cm,rotate=170] at (1.5,-3) (ac2) {};
\node[circle,draw] at ($(ac2)+(-0.7,-0.7)$) {$i_2$};
\node [aircraft top,fill=black,minimum width=1cm,rotate=80] at (-5,0.9) (ac3) {};
\node[circle,draw] at ($(ac3)+(0.1,1.4)$) {$i_3$};

\draw[dashed] (ac1) .. controls ($(ac1)-(-28:5)$) and (-2,5) .. (-6,5);
\draw[dashed] (ac3) .. controls ($(ac3)-(80:3)$) and (-6,-3) .. (-4,-5);
\draw[dashed] (ac2) .. controls ($(ac2)-(167:5)$) and (4.5,3)  .. (4.5,5);

\node[] at (-0.5,1.0) {$\Omega_\varepsilon$};
\end{tikzpicture}
				\caption{\boldmath Arrival reward.}
				\label{fig:arr_reward}
			\end{subfigure}\hspace{0.03\textwidth}
			\begin{subfigure}[c]{0.455\textwidth}
				\centering
				\begin{tikzpicture}[scale=0.65]
\pgfuseshading{cav_reward_shade}

\fill[opacity=0] (0cm,0cm) circle (5cm);
\draw[orange,thick] (0cm,0cm) circle (4cm);
\draw[red!60!black,thick] (0cm,0cm) circle (2.5cm);
\node [aircraft top,fill=black,minimum width=1cm,rotate=0] at (0,0) (ac0) {};
\node[draw,circle,minimum height=0.8cm] at (0.7cm,-0.9cm) (label_i) {$i$};

\draw[thick,-latex'] (0cm,0cm) -- node[pos=0.5,above,rotate=-10]{$\varepsilon_{CAV}$}($(0,0)+(-10:2.5)$);
\draw[thick,-latex'] (0cm,0cm) -- node[pos=0.8,above,rotate=50]{$\delta_{CAV}$}($(0,0)+(50:4)$);

\node [aircraft top,fill=black,minimum width=1cm,rotate=-20] at (-0.2,-1.8) (ac1) {};
\node[circle,draw,minimum height=0.8cm] at ($(ac1)+(0.4,-1.4)$) {$j_1$};
\node [aircraft top,fill=black,minimum width=1cm,rotate=140] at (4,3) (ac3) {};
\node[circle,draw,minimum height=0.8cm] at ($(ac3)+(0.1,1)$) {$j_3$};
\node [aircraft top,fill=black,minimum width=1cm,rotate=55] at (-2.7,1.8) (ac2) {};
\node[circle,draw] at ($(ac2)+(1.0,1.0)$) {$j_2$};

\draw[dashed] (ac1) .. controls ($(ac1)-(-20:5)$) and (-5.5,-3) .. (-5.5,-5);
\draw[dashed] (ac2) .. controls ($(ac2)-(57:5)$) and (-4.5,2.5) .. (-5.5,3.5);
\draw[dashed] (ac3) .. controls ($(ac3)-(145:3)$) and (5,-3) .. (5.5,-5);

\node[] at (-0.8,1) {$\Omega_{\varepsilon,i}$};
\node[] at (0.3,3.15) {$\Omega_{\delta,i}$};
\end{tikzpicture}
				\caption{\boldmath Collision avoidance reward. }
				\label{fig:cav_reward}
			\end{subfigure}
			\caption{Different regions and scenarios for arrival reward (Fig. \ref{fig:arr_reward}) and collision avoidance reward (Fig. \ref{fig:cav_reward}).}
			
		\end{figure}
		Fig. \ref{fig:cav_reward} showcases possible constellations for different collision avoidance rewards. $j_1$ is located inside the $\varepsilon$-region of $i$, therefore both agents are considered to be part of a collision. The maximum negative reward is awarded to both. $j_2$ is located inside the $\delta$-region which is tolerated (with a smaller negative reward), but not preferred. $j_3$ and $i$ are clear of conflict.
		\item \textbf{Time minimization for arrival at target position:}
		\par The time reward $R_{TME}$ is realized by awarding a small negative reward for each time step. However, due to different distances between the individual agents' original locations and destinations, the negative reward per time step is normalized by the Euclidean distance between the origin and destination for each agent,so that the total time reward $\mathcal{R}_{TME}$ is equal for all agents, independent of the length of their route.
		\item \textbf{Acceleration minimization:}
		\par For the acceleration reward $R_{ACC}$, the absolute value of the agent's acceleration is integrated over time and the sign is inverted to be coherent with the following definition of the reward:
		\begin{equation}
			R_{ACC}=-\int\limits_t\left|\bm{a}\right|\text{d}t
		\end{equation}
	\end{enumerate}
	\par The total reward is the weighted sum of the individual rewards. The weights can be determined and adjusted according to the behavior of the agent. In the context of a homogeneous multi-agent system, i.e., all agents follow use the same policy, each agents' experience is stacked and treated as multiple episodes of a single agent.
	
	\section{Path Planning for Variably-Sized Observation Spaces}
	Dealing with observation spaces of changing size constitutes an area of special interest and provides challenges since artificial neural networks commonly take a constant amount of data as input. Therefore, some kind of encoding mechanism is required. Unlike other approaches that use occupancy maps, the proposed approach incorporates the encoding problem into the path planning problem by exploring the use of Long Short-Term Memory for the purpose of encoding.
	\subsection{Long Short-Term Memory and Policy Network Architecture}
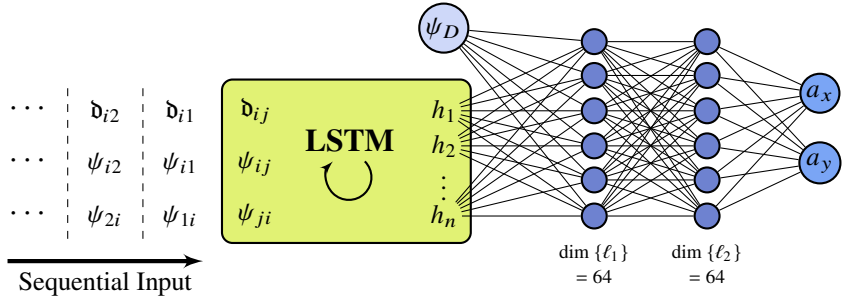
\begin{figure}[t!]
	\centering
	\begin{tikzpicture}

\node[] at (-1,0) (ldot1) {\large$\cdots$};
\node[] at (-1,-0.7) (ldot2) {\large$\cdots$};
\node[] at (-1,-1.4) (ldot3) {\large$\cdots$};

\node[] at (0,0) (di2) {$\mathfrak{d}_{i2}$};
\node[] at (0,-0.7) (psii2) {$\psi_{i2}$};
\node[] at (0,-1.4) (psi2i) {$\psi_{2i}$};

\node[] at (1,0) (di1) {$\mathfrak{d}_{i1}$};
\node[] at (1,-0.7) (psii1) {$\psi_{i1}$};
\node[] at (1,-1.4) (psi1i) {$\psi_{1i}$};

\draw[dashed] (-0.5,0.3) -- (-0.5,-1.75);
\draw[dashed] (0.5,0.3) -- (0.5,-1.75);

\node[] at (2,0) (dij) {$\mathfrak{d}_{ij}$};
\node[] at (2,-0.7) (psiij) {$\psi_{ij}$};
\node[] at (2,-1.4) (psiji) {$\psi_{ji}$};

\node[circle,inner sep=0pt] at (4.5,0) (h1) {$h_1$};
\node[circle,inner sep=0pt] at (4.5,-0.4666) (h2) {$h_2$};
\node[circle,inner sep=0pt] at (4.5,-0.92) (h3) {\vdots};
\node[circle,inner sep=0pt] at (4.5,-1.40) (h4) {$h_n$};
\node[] at (3.25,-0.4) (lstm) {\large \textbf{LSTM}};
\draw[thick,-latex'] ($(lstm.south)+(0.2,0)$) arc(45:-225:0.3);

\begin{scope}[on background layer]
\node[draw,rounded corners,fit = (dij) (h4),fill={rgb,255:red,225; green,242; blue,116},thick] (block_lstm){};
\end{scope}

\node[draw,circle,fill={rgb,255:red,110; green,130; blue,208},thick] at (6.5,0) (l21) {};
\node[draw,circle,fill={rgb,255:red,110; green,130; blue,208},thick] at (6.5,-0.46666) (l22) {};
\node[draw,circle,fill={rgb,255:red,110; green,130; blue,208},thick] at (6.5,-0.92) (l23) {};
\node[draw,circle,fill={rgb,255:red,110; green,130; blue,208},thick] at (6.5,-1.4) (l24) {};
\node[draw,circle,fill={rgb,255:red,110; green,130; blue,208},thick] at (6.5,0.4699) (l25) {};
\node[draw,circle,fill={rgb,255:red,110; green,130; blue,208},thick] at (6.5,0.92) (l26) {};
\node[anchor=north] at (6.5,-1.6) (descrpit1) {\scriptsize$\begin{matrix}
	\dim\left\{\ell_1\right\}\\
	=64
	\end{matrix}$};

\node[draw,circle,text=black,fill={rgb,255:red,206; green,216; blue,249},inner sep=1pt,thick] at (4.5,1.1) (sarr) {$\psi_D$};

\draw[thin] (sarr) -- (l21);
\draw[thin] (sarr) -- (l22);
\draw[thin] (sarr) -- (l23);
\draw[thin] (sarr) -- (l24);
\draw[thin] (sarr) -- (l25);
\draw[thin] (sarr) -- (l26);

\draw[thin] (h1) -- (l21);
\draw[thin] (h1) -- (l22);
\draw[thin] (h1) -- (l23);
\draw[thin] (h1) -- (l24);
\draw[thin] (h1) -- (l25);
\draw[thin] (h1) -- (l26);

\draw[thin] (h2) -- (l21);
\draw[thin] (h2) -- (l22);
\draw[thin] (h2) -- (l23);
\draw[thin] (h2) -- (l24);
\draw[thin] (h2) -- (l25);
\draw[thin] (h2) -- (l26);

\draw[thin] (h4) -- (l21);
\draw[thin] (h4) -- (l22);
\draw[thin] (h4) -- (l23);
\draw[thin] (h4) -- (l24);
\draw[thin] (h4) -- (l25);
\draw[thin] (h4) -- (l26);

\node[draw,circle,fill={rgb,255:red,110; green,130; blue,208},thick] at (8,0) (l31) {};
\node[draw,circle,fill={rgb,255:red,110; green,130; blue,208},thick] at (8,-0.46666) (l32) {};
\node[draw,circle,fill={rgb,255:red,110; green,130; blue,208},thick] at (8,-0.92) (l33) {};
\node[draw,circle,fill={rgb,255:red,110; green,130; blue,208},thick] at (8,-1.4) (l34) {};
\node[draw,circle,fill={rgb,255:red,110; green,130; blue,208},thick] at (8,0.4699) (l35) {};
\node[draw,circle,fill={rgb,255:red,110; green,130; blue,208},thick] at (8,0.92) (l36) {};
\node[anchor=north] at (8,-1.6) (descrpit2) {\scriptsize$\begin{matrix}
	\dim\left\{\ell_2\right\}\\
	=64
	\end{matrix}$};

\draw[thin] (l21) -- (l31);
\draw[thin] (l21) -- (l32);
\draw[thin] (l21) -- (l33);
\draw[thin] (l21) -- (l34);
\draw[thin] (l21) -- (l35);
\draw[thin] (l21) -- (l36);

\draw[thin] (l22) -- (l31);
\draw[thin] (l22) -- (l32);
\draw[thin] (l22) -- (l33);
\draw[thin] (l22) -- (l34);
\draw[thin] (l22) -- (l35);
\draw[thin] (l22) -- (l36);

\draw[thin] (l23) -- (l31);
\draw[thin] (l23) -- (l32);
\draw[thin] (l23) -- (l33);
\draw[thin] (l23) -- (l34);
\draw[thin] (l23) -- (l35);
\draw[thin] (l23) -- (l36);

\draw[thin] (l24) -- (l31);
\draw[thin] (l24) -- (l32);
\draw[thin] (l24) -- (l33);
\draw[thin] (l24) -- (l34);
\draw[thin] (l24) -- (l35);
\draw[thin] (l24) -- (l36);

\draw[thin] (l25) -- (l31);
\draw[thin] (l25) -- (l32);
\draw[thin] (l25) -- (l33);
\draw[thin] (l25) -- (l34);
\draw[thin] (l25) -- (l35);
\draw[thin] (l25) -- (l36);

\draw[thin] (l26) -- (l31);
\draw[thin] (l26) -- (l32);
\draw[thin] (l26) -- (l33);
\draw[thin] (l26) -- (l34);
\draw[thin] (l26) -- (l35);
\draw[thin] (l26) -- (l36);

\node[draw,circle,fill={rgb,255:red,122; green,165; blue,245},thick,inner sep=1pt] at (9.5,0.2333) (ax) {$a_x$};
\node[draw,circle,fill={rgb,255:red,122; green,165; blue,245},thick,inner sep=1pt] at (9.5,-0.693) (ay) {$a_y$};

\draw[thin] (l31) -- (ax);
\draw[thin] (l32) -- (ax);
\draw[thin] (l33) -- (ax);
\draw[thin] (l34) -- (ax);
\draw[thin] (l35) -- (ax);
\draw[thin] (l36) -- (ax);

\draw[thin] (l31) -- (ay);
\draw[thin] (l32) -- (ay);
\draw[thin] (l33) -- (ay);
\draw[thin] (l34) -- (ay);
\draw[thin] (l35) -- (ay);
\draw[thin] (l36) -- (ay);

\draw[ultra thick,-latex'] (-1.3,-2)--node[pos=0.5,below]{Sequential Input}(1.3,-2); 

\end{tikzpicture}
	\caption{Layout of the policy network with a leading LSTM, a trailing fully connected neural network, and actions in Cartesian coordinates as the outputs.}
	\label{fig:policy_net_lstm}
\end{figure}

	A Long Short-Term Memory is an advanced type of recurrent structure in the form of a neural network that is mainly used for time-series applications \cite{Hochreiter.1997,Gers.1999}. The input of the structure is denoted as $\bm{x}\in\mathbb{R}^m$ and the output, commonly referred to as the hidden layer\footnote{It shall be noted that, despite the implied connotation, the term \textit{hidden layer} refers to the output of the LSTM and not the hidden layers in a feed-forward ANN.}, as $\bm{h}\in\mathbb{R}^n$. Internally, the LSTM is composed of three gates: the input gate $\bm{i}$, the forget gate $\bm{f}$, and the output gate $\bm{o}$, each dependent on the values of the input and hidden layers. The core of the LSTM module is the cell $\bm{c}$, which can be regarded as the organizational center of the module. The input and forget gates regulate what information will be stored in the cell for the next update cycle based on the current input, cell, and hidden states. The output gate is responsible for regulating the information that is transferred from the cell to the hidden state based on the current input and hidden states. The data is passed sequentially into the LSTM. In the presented case, each tuple $\bm{x} = \begin{bmatrix}
		\mathfrak{d}_{ij}&\psi_{ij}&\psi_{ji}
	\end{bmatrix}^T$, identifying a vehicle's relative position and heading (see Fig. \ref{fig:relative_angles}), is taken as an element $t$ of the input sequence into the LSTM. For a scenario with $\dim\left\{\mathcal{V}\right\}$ vehicles, the LSTM has to be evaluated $\dim\left\{\mathcal{V}\right\}-1$ times. Only the hidden states after the last iteration are used for further calculations as illustrated in Fig. \ref{fig:policy_net_lstm}. The following equations describe the behavior of the LSTM:
	\begin{align}
		\bm{i}_t&=\sigma\underbrace{\left(\bm{W}_i\cdot\bm{x}_t+\bm{U}_i\cdot\bm{h}_{t-1}+\bm{b}_i\right)}_{\bm{I}_t}\label{eq:i_gate}\\
		\bm{f}_t&=\sigma\underbrace{\left(\bm{W}_f\cdot\bm{x}_t+\bm{U}_f\cdot\bm{h}_{t-1}+\bm{b}_f\right)}_{\bm{F}_t}\label{eq:f_gate}\\
		\bm{o}_t&=\sigma\underbrace{\left(\bm{W}_o\cdot\bm{x}_t+\bm{U}_o\cdot\bm{h}_{t-1}+\bm{b}_o\right)}_{\bm{O}_t}\label{eq:o_gate}\\
		\bm{c}_t&= \bm{f}_t\circ \bm{c}_{t-1}+\bm{i}_t\circ\underbrace{\tanh\underbrace{\left(\bm{W}_c\cdot\bm{x}_t+\bm{U}_c\cdot\bm{h}_{t-1}+\bm{b}_c\right)}_{\bm{G}_t}}_{\bm{g}_t}\label{eq:cell}\\
		\bm{h}_t&=\bm{o}_t\circ\tanh\left(\bm{c}_t\right)\label{eq:hidden_layer}
	\end{align}
	\par
\begin{figure}[b!]
		\centering
		\begin{tikzpicture}
\node[draw,circle,minimum height=1cm,thick,fill={rgb,255:red,173; green,221; blue,240}] at (0,0) (x) {$\boldsymbol{x}$};
\node[draw,circle,minimum height=1cm,thick,fill={rgb,255:red,110; green,130; blue,208}] at (7cm,0) (c) {$\boldsymbol{c}$};
\node[draw,circle,minimum height=1cm,thick,fill={rgb,255:red,122; green,165; blue,245}] at (12cm,0) (h) {$\boldsymbol{h}$};
\node[draw,circle,minimum height=1cm,thick,fill={rgb,255:red,206; green,216; blue,249}] at (3cm,3.5) (i) {$\boldsymbol{i}$};
\node[draw,circle,minimum height=1cm,thick,fill={rgb,255:red,206; green,216; blue,249}] at (7cm,3.5) (f) {$\boldsymbol{f}$};
\node[draw,circle,minimum height=1cm,thick,fill={rgb,255:red,206; green,216; blue,249}] at (11cm,3.5) (o) {$\boldsymbol{o}$};

\node[inner sep=0.3pt] at ($(i.west)-(0.7,0)$) (sig1) {$\sigma$};
\draw[-latex'] (sig1) -- (i);
\node[inner sep=0pt] at ($(sig1.west)-(0.5,0)$) (sum1) {\summation};
\draw[-latex'] (sum1) -- (sig1);
\draw[-latex'] ($(sum1.west)-(0.7,0)$) -- node[pos=0,anchor=south west]{$b_i$} (sum1);
\node[inner sep = 0.3pt] at ($(sum1.north)+(0,0.7)$) (wi) {$W_i$};
\draw[-latex'] (wi) -- (sum1);
\node[inner sep = 0.3pt] at ($(sum1.south)-(0,0.7)$) (ui) {$U_i$};
\draw[-latex'] (ui) -- (sum1);

\node[inner sep=0.3pt] at ($(f.west)-(0.7,0)$) (sig2) {$\sigma$};
\draw[-latex'] (sig2) -- (f);
\node[inner sep=0pt] at ($(sig2.west)-(0.5,0)$) (sum2) {\summation};
\draw[-latex'] (sum2) -- (sig2);
\draw[-latex'] ($(sum2.west)-(0.7,0)$) -- node[pos=0,anchor=south west]{$b_f$} (sum2);
\node[inner sep = 0.3pt] at ($(sum2.north)+(0,0.7)$) (wf) {$W_f$};
\draw[-latex'] (wf) -- (sum2);
\node[inner sep = 0.3pt] at ($(sum2.south)-(0,0.7)$) (uf) {$U_f$};
\draw[-latex'] (uf) -- (sum2);

\node[inner sep=0.3pt] at ($(o.west)-(0.7,0)$) (sig3) {$\sigma$};
\draw[-latex'] (sig3) -- (o);
\node[inner sep=0pt] at ($(sig3.west)-(0.5,0)$) (sum3) {\summation};
\draw[-latex'] (sum3) -- (sig3);
\draw[-latex'] ($(sum3.west)-(0.7,0)$) -- node[pos=0,anchor=south west]{$b_o$} (sum3);
\node[inner sep = 0.3pt] at ($(sum3.north)+(0,0.7)$) (wo) {$W_o$};
\draw[-latex'] (wo) -- (sum3);
\node[inner sep = 0.3pt] at ($(sum3.south)-(0,0.7)$) (uo) {$U_o$};
\draw[-latex'] (uo) -- (sum3);

\draw[-latex'] (x) |- ($(wi)+(0,0.7)$) -- (wi);
\draw[-latex'] ($(wi)+(0,0.7)$) -| (wf);
\draw[-latex'] ($(wf)+(0,0.7)$) -| (wo);

\draw[-latex'] (h) |- ($(uo)-(0,0.7)$) -- (uo);
\draw[-latex'] ($(uo)-(0,0.7)$)-|(uf);
\draw[-latex'] ($(uf)-(0,0.7)$)-|(ui);

\node[inner sep = 0pt] at ($(c.west)-(0.7,0)$) (csum) {\summation};
\draw[-latex'] (csum) -- (c);
\node[inner sep = 0pt] at ($(csum)-(1,0)$) (had1) {\hadamard};
\draw[-latex'] (had1) -- (csum);

\node[inner sep=0.3pt] at ($(had1.west)-(0.9,0)$) (tanh1) {$\tanh$};
\draw[-latex'] (tanh1) -- (had1);
\node[inner sep=0pt] at ($(tanh1.west)-(0.7,0)$) (sum4) {\summation};
\draw[-latex'] (sum4) -- (tanh1);
\draw[-latex'] ($(sum4.west)-(0.7,0)$) -- node[pos=0,anchor=south west]{$b_c$} (sum4);
\node[inner sep = 0.3pt] at ($(sum4.north)+(0,0.7)$) (uc) {$U_c$};
\draw[-latex'] (uc) -- (sum4);
\node[inner sep = 0.3pt] at ($(sum4.south)-(0,0.7)$) (wc) {$W_c$};
\draw[-latex'] (wc) -- (sum4);

\draw[-latex'] ($(uf)-(2.08,0.7)$)-|(uc);
\draw[-latex'] ($(x)+(0,1)$)-|($(wc)-(1.5,0)$)--(wc);
\draw[-latex'] (i) to[in=90,out=-45] (had1);

\node[inner sep=0pt] at ($(csum)+(0.5,1.3)$) (had2) {\hadamard};
\draw[-latex'] (had2) to[in=90, out=-140] (csum);
\draw[-latex'] (c) to[in=-20, out=80] (had2);
\draw[-latex'] (f) to[in=90,out=-120] (had2);

\node[inner sep = 0.3pt] at ($(c.east)+(1,0)$) (tanh2) {$\tanh$};
\draw[-latex'] (c) -- (tanh2);
\node[inner sep=0pt] at ($(tanh2)+(1.5,0)$) (had3) {\hadamard};
\draw[-latex'] (tanh2) -- (had3);
\draw[-latex'] (had3) -- (h);
\draw[-latex'] (o) to[in=90,out=-120] (had3);
\end{tikzpicture}
		\caption{Structure of an LSTM module.}
		\label{fig:lstm}
	\end{figure}
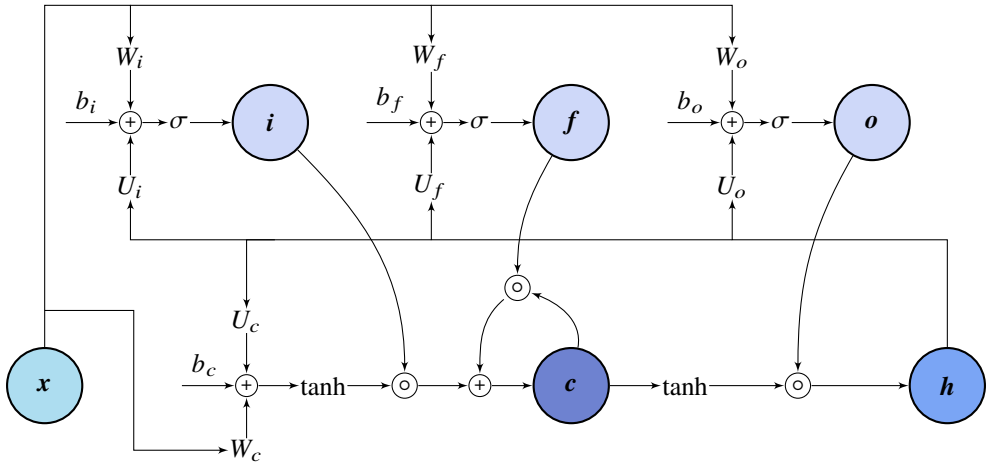

 Throughout Eqs. (\ref{eq:i_gate}) to (\ref{eq:hidden_layer}), $\sigma$ denotes the sigmoid function and $\circ$ denotes the Hadamard Product. $\bm{b}$ always represents a constant bias vector of size $\mathbb{R}^n$. $\bm{W}$ and $\bm{U}$ are weight matrices of dimensions $\mathbb{R}^{n\times m}$ and $\mathbb{R}^{n\times n}$, respectively. Hence, $\bm{i},\bm{f},\bm{o},\bm{c}\in\mathbb{R}^n$. Fig. \ref{fig:lstm} illustrates the architecture of an LSTM module.
	\par In a larger context, the LSTM mainly contributes to the objective of collision avoidance from the list of desirable path planning properties. To guide the vehicle to its destination, the relative angle between the current heading and the destination, denoted as $\psi_D$, is necessary. The hidden states after the $\left(\dim\left\{\mathcal{V}\right\}-1\right)$-th iteration represent an encoded version of the positions and heading of all vehicles. This is equivalent to the commonly used occupation map $\mathcal{S}_{map}$, where the positions of vehicles are marked. Using LSTMs, the relative heading information of $j$, $\psi_{ji}$, could be included as well without increasing the state space significantly. Generally, $\dim\left\{\bm{h}\right\}\ll\dim\left\{\mathcal{S}_{map}\right\}$.
	\par The size of $\dim\left\{\bm{h}\right\}$ is a trade-off between more information to transmit and a smaller number of parameters to estimate. We chose $\dim\left\{\bm{h}\right\}=63$ based on a simple estimate using the entropy of an occupancy map by treating the map as discrete random variable \cite{Shannon.1948}. The hidden states after the last iteration of the LSTM are then stacked with $\psi_D$ and used as input to a fully connected neural network with two hidden layers with 64 neurons each, using the hyperbolic tangent as activation function. The last layer, the output layer, consists of $dim\left\{\Omega\right\}$ neurons, where $dim\left\{\Omega\right\}$ denotes the number of spatial dimensions. The complete layout is depicted in Fig. \ref{fig:policy_net_lstm}.
	\subsection{Scenario Creation for Efficient Training}
	\par Applying the proposed method to use LSTMs as an encoding mechanism, thereby enabling work with observation spaces of variable size without the prior use of encoding mechanisms like maps, raises the problem of efficient scenario creation.
	\par Compared to commonly used encoding mechanisms for position encoding, LSTMs require more specific training. Using occupation maps as an encoded representation of the agents' relative positions, the task can be treated as interpreting an image and calculating the best commands from it. The structure of the neural network presented in this work incorporates the encoding process; thus, we must not only learn the transformation between encoded versions of the physical state into commands, but also the encoding itself.
	\par Regarding the task, only the collision avoidance reward benefits from complex scenarios, whereas an optimal behavior with respect to arrival at destination, time optimality, and minimization of acceleration does not require a large number of agents. On the contrary, large quantities of agents, especially in an initial phase of the training, tend to impact the agent's performance negatively due to scenarios where the probability of a positive outcome approaches zero. Therefore, it is advised to sample the amount of agents $\dim\left\{\mathcal{V}\right\}$ from a geometric probability distribution:
	\begin{equation}
		\dim\left\{\mathcal{V}\right\}\sim\left(1-p\right)^{n-1}\cdot p\label{eq:geo_distribution}
	\end{equation}
	where $p\in\left(0,1\right)$ is a constant which is suggested to be set to $p=1/3$. Using these parameters, unnecessarily computation-intensive calculations can be avoided and the training becomes more efficient.

	\section{Results}
	\subsection{Overall Performance}\label{chap:overall_path_planning}
	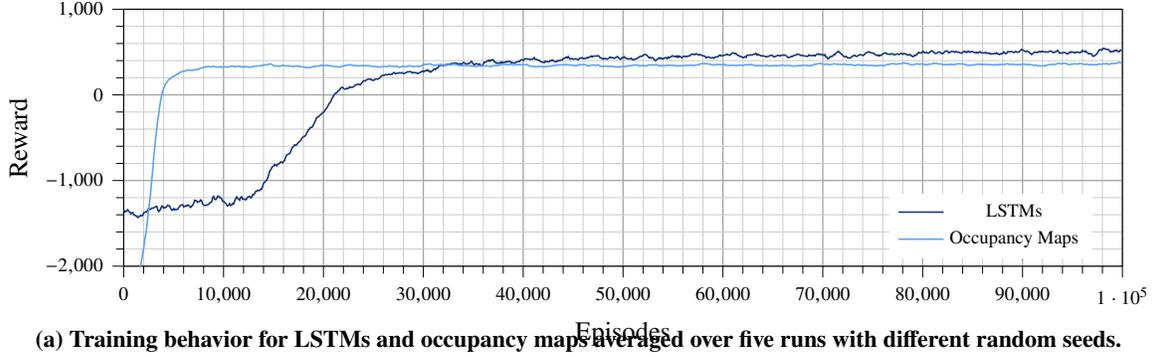
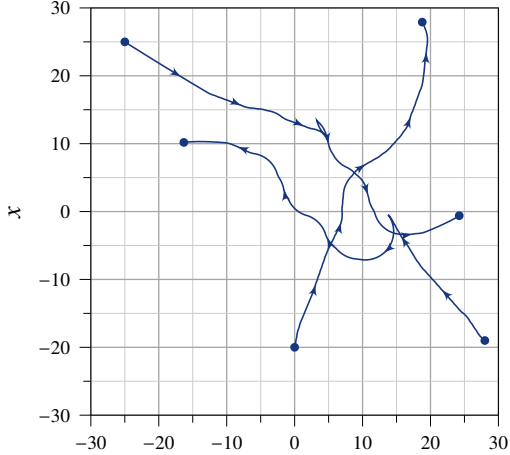
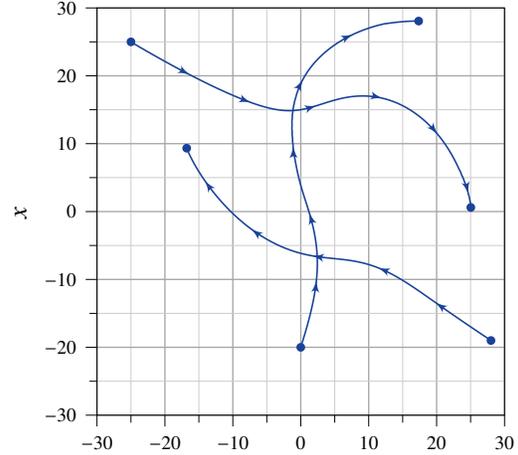
\begin{figure}[t!]
		
		\centering
		\begin{subfigure}[b]{\textwidth}
			\centering
			\begin{tikzpicture}[fill between/on layer=axis background]

\definecolor{color0}{rgb}{0,0,0}
\definecolor{color0b}{rgb}{0.5,0.5,0.5}
\definecolor{color1}{rgb}{0.1960784313,0.309803921,0.6666666}
\definecolor{color1b}{RGB}{22,55,128}
\definecolor{color2}{rgb}{0.1843137,0.68627450,0.866666666}
\definecolor{color3}{rgb}{0,0.541176,0}
\definecolor{color3b}{RGB}{92,162,247}
\definecolor{color4}{rgb}{0.57254,0.82745098,0.058823529}
\definecolor{color5}{rgb}{0.96078431372,0.32941176470,0.074509803}
\definecolor{color6}{rgb}{0.99607843,0.69411764,0}
\definecolor{color7}{rgb}{0.952941176,0.30980392,0.541176470}
\definecolor{color8}{rgb}{0.69019607,0,0.69019607}
\begin{axis}[
width=0.9\textwidth,
height=5cm,
tick align=outside,
tick pos=left,
grid style={line width=.25pt, draw=gray!40},
major grid style={line width=.5pt,draw=gray!70},
minor tick num=4,
xlabel={Episodes},
ylabel={Reward},
xtick={0,10000,20000,30000,40000,50000,60000,70000,80000,90000,100000},
ytick={-3000,-2000,-1000,0,1000},
xtick style={color=black},
xmin=0,xmax=100000,
ymin=-2000, ymax=1000,
ytick style={color=black},
xmajorgrids,
xminorgrids,
ymajorgrids,
yminorgrids,
scaled x ticks=false,
ticklabel style = {font=\scriptsize},
ticklabel style = {font=\scriptsize},
legend style={draw=none,font=\scriptsize,cells={align=left}},
legend pos = south east
]

\addplot [semithick,name path=mean,color1b]
table {training_sim_id_24.txt};
\addlegendentry{LSTMs}

\addplot [semithick,name path=mean,color3b]
table {training_sim_id_20.txt};
\addlegendentry{Occupancy Maps}

\end{axis}

\end{tikzpicture}
			\captionsetup{skip=-8mm}
			\caption{Training behavior for LSTMs and occupancy maps averaged over five runs with different random seeds.}
			\label{fig:opp_a}
		\end{subfigure}
		\vspace{7mm}
		\begin{subfigure}[c]{0.49\textwidth}
			\centering
			\include{trajec_sim_id_20}
			\captionsetup{skip=-8mm}
			\caption{Ground tracks for three agents using occupancy maps.}
			\label{fig:opp_b}
		\end{subfigure}
		\begin{subfigure}[c]{0.49\textwidth}
			\centering
			\include{trajec_sim_id_22}
			\captionsetup{skip=-8mm}
			\caption{Ground tracks for three agents using LSTMs.}
			\label{fig:opp_c}
		\end{subfigure}
		\vskip -5mm
		\caption{Training and ground tracks for three agents trained for 100,000 episodes using occupancy maps and LSTMs.}
		\label{fig:overall_path_planning}
	\end{figure}
	As depicted in Fig. \ref{fig:overall_path_planning}, there are significant differences when comparing occupancy maps, the method currently used in many applications, to LSTMs as an encoding mechanism. As indicated in Fig. \ref{fig:opp_a}, the policy using a occupancy map of size $8\times 25$ (8 bins and 25 bins for $\mathfrak{d}_{ij}$ and $\psi_{ij}$, respectively) on average converges faster than the LSTM-based policy. However, the LSTM-based policy collects a higher maximum average reward during training. The difference between the two settings becomes even more obvious when reviewing ground tracks in Fig. \ref{fig:opp_b} and Fig. \ref{fig:opp_c}, respectively. The use of occupancy maps results in unwanted trajectories with sharp turns and high accelerations. In comparison, using LSTMs results in smooth trajectories, satisfying all conditions, after the same number of training episodes. 
	
	\subsection{Collision Avoidance}\label{chap:collision_avoidance_results}
	\par
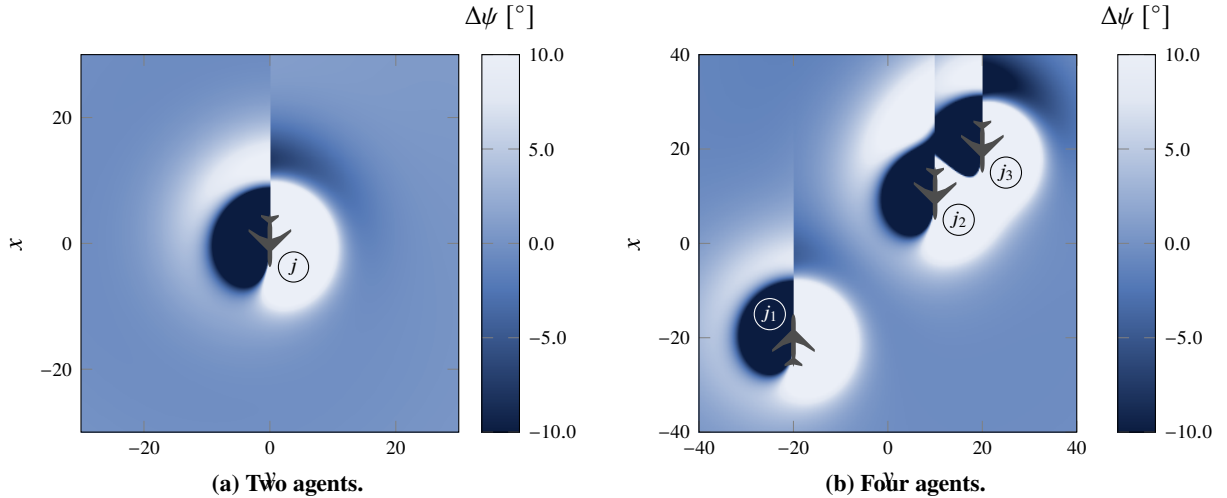
\begin{figure}[b!]
		\centering
		\begin{subfigure}[c]{3.0in}
			\centering
			\begin{tikzpicture}

 \begin{axis}[
            view={0}{90},   
            xlabel=$y$,
            ylabel=$x$,
            colormap name={custom2c},
            colorbar,
            colorbar style={
                title=$\Delta\psi\left[^\circ\right]$,
                yticklabel style={
                    /pgf/number format/.cd,
                    fixed,
                    precision=1,
                    fixed zerofill,
                },
            },
        	ticklabel style={font=\scriptsize},
            enlargelimits=false,
            axis on top,
            point meta min=-10,
            point meta max=10,
            width=2.6in,
            height=2.6in
            %
        ]
\addplot graphics[xmin=-30,ymin=-30,xmax=30,ymax=30] {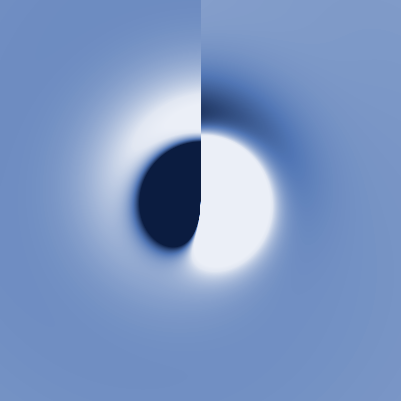};
\node [aircraft top,fill=black!70,minimum width=0.7cm,rotate=-90] at (0,0) (i) {};
\node[circle,draw,inner sep=1.5pt] at (3.75,-3.75) {\scriptsize$j$};
\end{axis}
\end{tikzpicture}
			\captionsetup{skip=-8mm}
			\caption{Two agents.}
			\label{fig:commanded2}
		\end{subfigure}\hspace{0.2in}
		\begin{subfigure}[c]{3.0in}
			\centering
			\begin{tikzpicture}
 \begin{axis}[
            view={0}{90},   
            xlabel=$y$,
            ylabel=$x$,
            colormap name={custom2c},
            colorbar,
            colorbar style={
                title=$\Delta\psi\left[^\circ\right]$,
                yticklabel style={
                    /pgf/number format/.cd,
                    fixed,
                    precision=1,
                    fixed zerofill,
                },
            },
        	ticklabel style={font=\scriptsize},
            enlargelimits=false,
            axis on top,
            point meta min=-10,
            point meta max=10,
            width=2.6in,
            height=2.6in
            %
        ]
\addplot graphics[xmin=-40,ymin=-40,xmax=40,ymax=40] {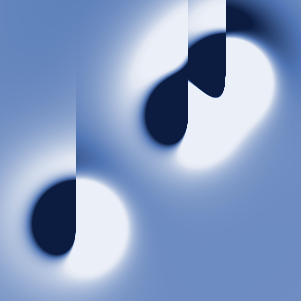};
\node [aircraft top,fill=black!70,minimum width=0.7cm,rotate=90] at (-20,-20) (i) {};
\node [aircraft top,fill=black!70,minimum width=0.7cm,rotate=-90] at (10,10) (i) {};
\node [aircraft top,fill=black!70,minimum width=0.7cm,rotate=-90] at (20,20) (i) {};
\node[circle,draw,inner sep=1pt,text=white,white] at (-25,-15) {\scriptsize$j_1$};
\node[circle,draw,inner sep=1pt] at (15,5) {\scriptsize$j_2$};
\node[circle,draw,inner sep=1pt] at (25,15) {\scriptsize$j_3$};
\end{axis}
\end{tikzpicture}
			\captionsetup{skip=-8mm}
			\caption{Four agents.}
			\label{fig:commanded5}
		\end{subfigure}
		
		\caption{\boldmath Commanded yaw angle for each position of the ego agent $\bm{i}$ relative to agent $\bm{j}$,$\bm{j_1}$,$\bm{j_2}$, and $\bm{j_3}$.}
		\label{fig:commanded_angles}
	\end{figure}

 Since collision avoidance is the essential objective, an evaluation will be shown to understand how the policy ensures collision avoidance. For the evaluation of $\dim\left\{\mathcal{V}\right\}$ agents, $\dim\left\{\mathcal{V}\right\}-1$ agents are placed at a static location and bearing in the phyiscal space. The remaining agent's position is altered to be at each position of the physical space once. The bearing is always oriented in the positive x-direction ($\psi_i=0$). For each location, the policy is evaluated and the commanded bearing difference $\Delta\psi$ is calculated. Fig. \ref{fig:commanded_angles} depicts a map, wherein the commanded angles are color-coded with Fig. \ref{fig:commanded2} showing a scenario for two agents, where the fixed agent is located at $\left(0,0\right)$ with $\psi_{j}=\pi$, causing a confrontational course. Fig. \ref{fig:commanded5} shows a more complex situation with four agents. Positions of the agents are: $\bm{r}_{j_1}=(-20,-20)$, $\bm{r}_{j_2}=(10,10)$, and $\bm{r}_{j_3}=(20,20)$. The bearings are: $\psi_{j_1}=0$, $\psi_{j_2}=\pi$, and $\psi_{j_3}=\pi$.
	\par Regarding the case of a two-agent scenario as depicted in Fig. \ref{fig:commanded2}, four observations can be made: 
	\begin{enumerate}
		\item In normal operations, the policy commands changes for collision avoidance early enough to prevent harsh corrections. This behavior is specifically influenced by $R_{ACC}$. However, in challenging situations like a trajectory crossing with a small included angle, it might not be possible to react as early as desired. Such cases are represented by larger actuator commands ($\varepsilon_{CAV}=3\text{m}$ and $\delta_{CAV}=10\text{m}$ for the example given). Therefore, the observable behavior of larger commanded changes in the bearing only appears when $\bm{r}\in\left(\Omega_{\varepsilon,CAV}\cup\Omega_{\delta,CAV}\right)$.
		\item Closer evaluation of the distribution of commands clearly displays that the presented policy favors right-handed evasion maneuvers. Even for positions located slightly to the left of the center axis, the policy commands a right turn. Since no particular requirements for this situation have been specified in the reward function, training with different random seeds also happens to result in policies favoring left-handed evasion maneuvers.
		\item In Fig. \ref{fig:commanded_angles}, an outer ring of counter actions can be observed outside the inner ring of actions. This behavior was not actively intended while designing the reward function; however, it leads to an interesting conclusion that the optimal behavior with respect to the reward function is the return to the original trajectory as intended before the evasive maneuver. Such a behavior enhances the predictibility of the whole system since the agents will, according to $R_{TME}$, choose a route that is time-optimal. The direct route is only interrupted by evasion maneuvers. 
		\item A non-omittable artifact is the immediate change in behavior between $\psi_{ij}=-\pi$ and $\psi_{ij}=\pi$. A model where $\psi_{ij}=-\pi$ and $\psi_{ij}=\pi$ not only have the same physical meaning, but also share the same quantitative description that would solve the problem. This could be achieved by using a two-dimensional vector space with one constraint for the description of $\psi_{ij}$. However, while testing the algorithm, no significant pitfalls were observed since the majority of all possible collision hazards were resolved early enough. 
	\end{enumerate}
	\par Regarding Fig. \ref{fig:commanded5}, for distant agents the behavior of the two agent-scenario is mostly shifted to the agent's position and stacked. This behavior can be observed for $j_1$, but not for the agents positioned closely to one another, $j_2$ and $j_3$. Close proximity of two agents increases the zone with larger commands beyond the size of $\Omega_{\delta,2}\cap\Omega_{\delta,1}$, i.e., keeping other agents away.
	\subsection{Scalability of the Policy}\label{chap:scalability}
	\begin{figure}[b!]
		\centering
		\input{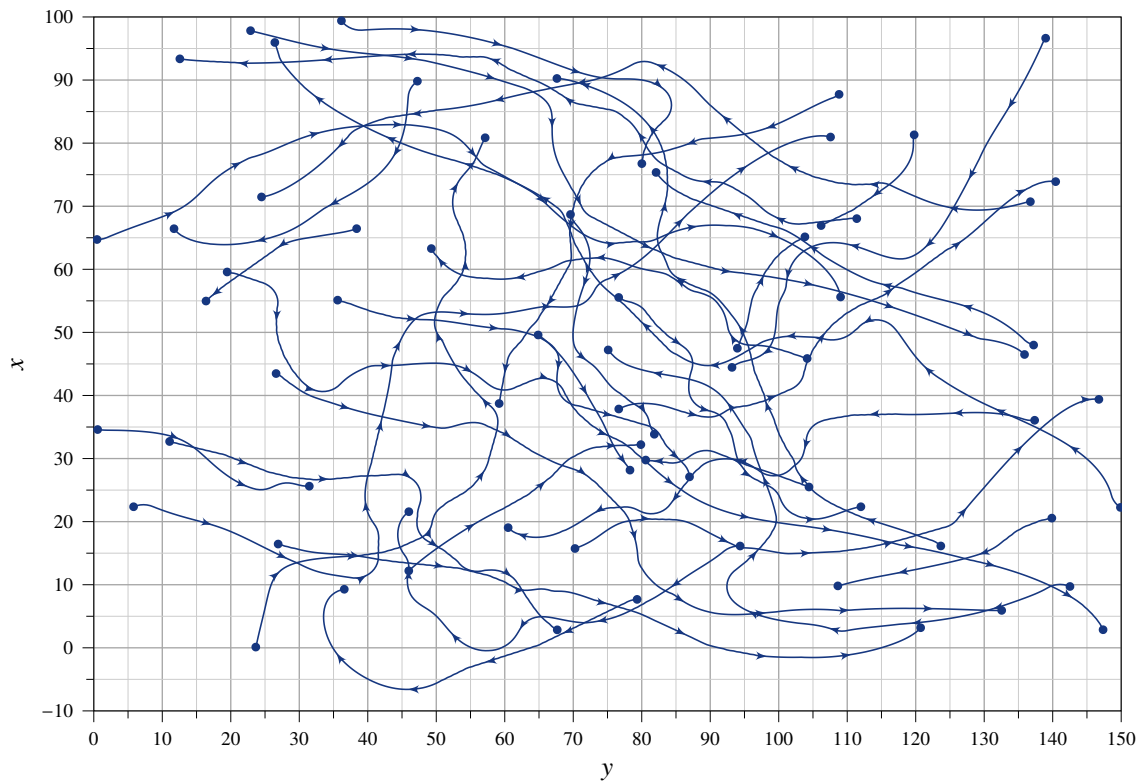}
		\caption{Ground tracks for 35 agents with flight direction and equal points in time \\indicated by arrows.}
		\label{fig:agent_8}
	\end{figure}
	\par Usually, control strategies are limited to the boundaries for which they were designed. Extrapolation is highly inadvisable since the control strategy has not been designed to handle such situations. However, the presented guidance strategy with LSTMs represents an exception. Even training with as few as five agents on limited physical space of $\dim\left\{\Omega\right\}=60~\text{m}\times60~\text{m}$, which keeps the required computations at a tolerable level and ensures fast initial success with regard to finding the target, leads to a policy that can be used for a seemingly indefinite number of agents as well as in a physical space that is larger than the training space by orders of magnitude.
	\par With an increasing number of agents, group dynamics become visible. Agents might block each other and prevent individuals from reaching their target position, especially in dense situations. The mutual blocking results in a formation of swarms. Swarms might be suboptimal for the individual agent; however, when considering all agents, the formation of swarms can be beneficial, at least in congested space.
	\par Fig. \ref{fig:agent_8} demonstrates the scalability of the policy. Ground tracks of 35 agents are shown, where the starting and terminating points have been randomly selected. The underlying policy was trained with $\dim\left\{\mathcal{V}\right\}\le 5$ on a constrained space of $60\text{m}\times 60\text{m}$. All agents reach their respective destinations while no collisions occur. Larger situations are possible, but ground tracks of such situations are vacuous. 
	
	\section{Complexity Analysis}\label{chap:complexity_analysis}
	\par Comparing methods such as MPC or RRTs to ANN-based path planning, the main difference is the time when the optimization takes place. A policy network can be trained for as long as necessary in order to estimate the solution of the optimization problem. However, while in evaluation mode, a non-recurrent neural network always has a computational complexity of $\mathcal{O}\left(1\right)$, rendering it perfect for real-time applications as the number of floating point operations required for one step is almost constant (slight variances may occur when estimating the values of non-rational activation functions). Adding the LSTM layer for the encoding process results in a change of computational complexity from constant to linear depending on the number of agents, i.e., $\mathcal{O}\left(\dim\left\{\mathcal{V}\right\}\right)$. This behavior is still better than other online methods, when limiting the maximum number of agents under consideration to a reasonable value, the method can be applied to real-time systems.
	\par Assuming a network architecture as depicted in Fig. \ref{fig:policy_net_lstm} with $w$ with w fully connected layers between the hidden state and the action layer and using the hyperbolic tangent as the activation function, the number of floating point operations can be calculated by:
	\begin{align}
		\text{FLOPs}&=\left(\dim\left\{\mathcal{V}\right\}-1\right)\cdot\left(8\cdot\dim\left\{\bm{h}\right\}^2+8\cdot\dim\left\{\bm{h}\right\}\cdot\dim\left\{\bm{x}\right\}+26\cdot\dim\left\{\bm{h}\right\}\right)\nonumber\\
		&+2\cdot\left(\dim\left\{\bm{h}\right\}+1\right)\cdot\dim\left\{\ell_1\right\}+4\cdot\dim\left\{\ell_1\right\}\nonumber\\
		&+\sum\limits_{k=1}^{w-1}\left(2\cdot\dim\left\{\ell_k\right\}\cdot\dim\left\{\ell_{k+1}\right\}+4\cdot\dim\left\{\ell_{k+1}\right\}\right)\nonumber\\
		&+2\cdot\dim\left\{\ell_w\right\}\cdot\dim\left\{\bm{a}\right\}+4\cdot\dim\left\{\bm{a}\right\}
	\end{align}
	For the model previously described (Fig. \ref{fig:policy_net_lstm}), this becomes: $\text{FLOPs}=\left(\dim\left\{\mathcal{V}\right\}-1\right)\cdot 34,902~\text{FLOPs}+17,160~\text{FLOPs}$. In the case of 20 agents, this results in $0.7152~\text{MFLOPs}$. That total number of floating point operations required to compute the control law enables every up-to-date processor to carry out these calculations multiple times per second. Even faster evaluation can be efficiently achieved  by reducing $\dim\left\{\bm{h}\right\}$, since the computational complexity scales quadratically with $\dim\left\{\bm{h}\right\}$, i.e., $\mathcal{O}\left(\dim\left\{\bm{h}\right\}^2\right)$.
		\section{Flight Tests}
		To validate the theoretically developed concepts, we present real-world flight tests with up to four simultaneously flying vehicles, demonstrating the feasibility of the LSTM-based encoding approach in combination with a guidance based on artificial neural networks in a real-world environment\footnote{For a video demonstration, please see {\footnotesize\url{https://schlichting.page.link/lstm_flight_test}}}.
		\subsection{Setup}
		\par Although the basic configuration for all four quadrotor drones is the same, there are three different frames utilized among the four drones, ranging in diameter from $0.20$ m to $0.48$ m. An off-the-shelf Pixhawk computer with PX4 as the framework can be found on all four drones \cite{Meier.2015}. The standard PX4 flight deck is augmented by a custom module dedicated to the evaluation of the policy and trained with the simplified simulation. To avoid connectivity problems, we prefer to have the policy evaluated onboard the main flight computer and not a companion computer, as often seen for applications involving the evaluation of artificial neural networks. For this purpose, the parameters of the trained policy are exported and the model is rewritten in C/C++ to integrate with PX4.
		\par
\begin{figure}[t!]
			\centering
			\begin{subfigure}[b]{3.2in}
				\centering
				\input{orsula.tex}
				\caption{Relevant components with mounted position.}
				\label{fig:orsula}
			\end{subfigure}
			\begin{subfigure}[b]{2.8in}
				\centering
				\includegraphics[width=3.0in]{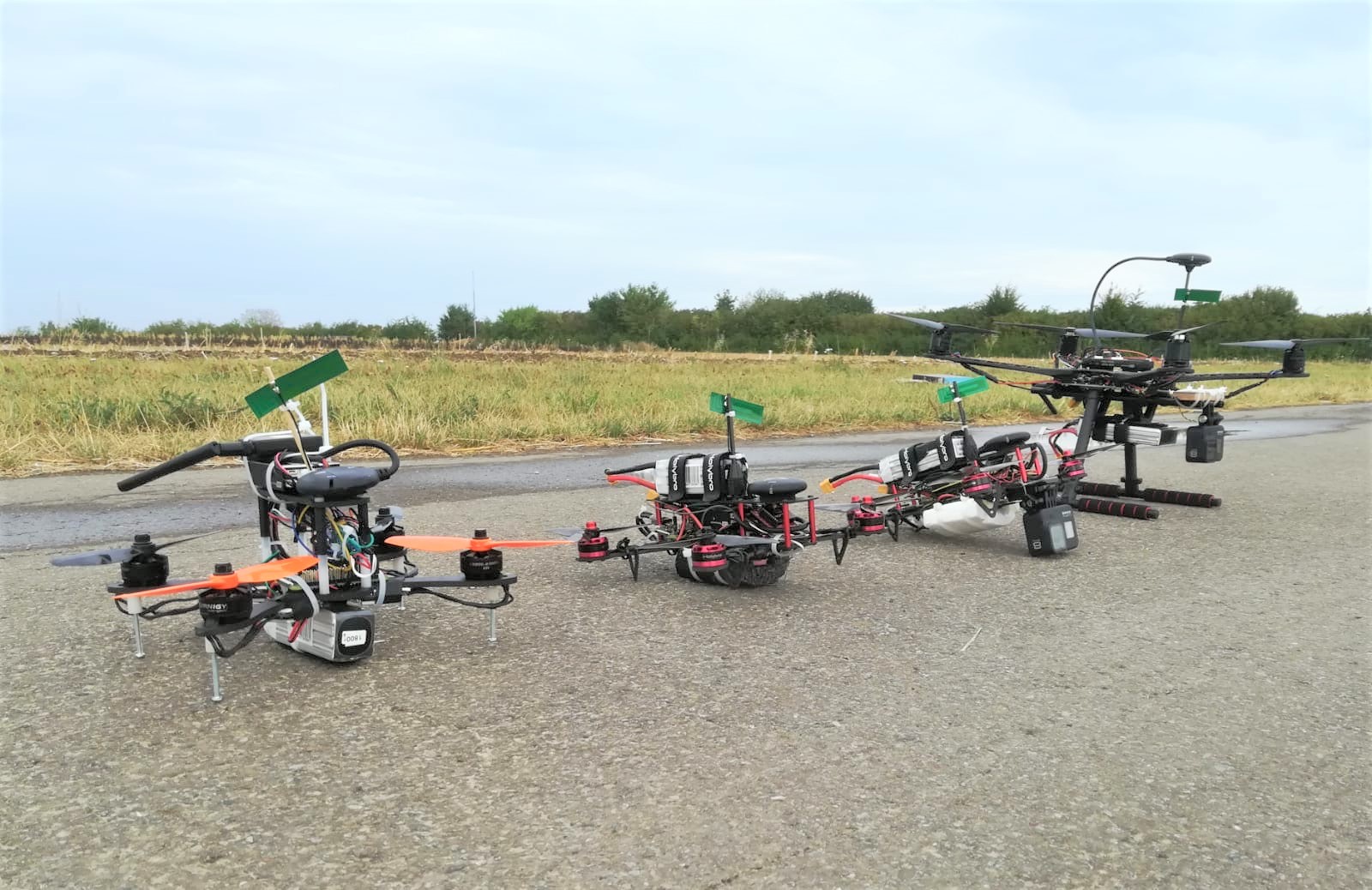}
				\caption{Drones used for flight training.}
				\label{fig:test_vehicles}
			\end{subfigure}
			\caption{Hardware used for test flights.}
			\label{fig:hardware}
		\end{figure}

\begin{figure}[t!]
			\centering
			\input{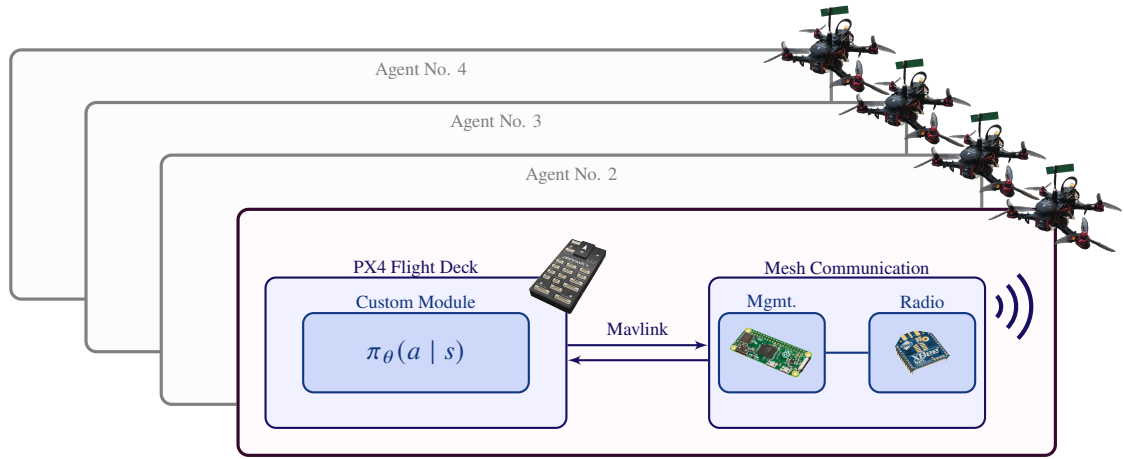}
			\caption{System schematics and implementation for inter-drone communication.}
			\label{fig:signal}
		\end{figure}

 As the relative positions between the vehicles are crucial to evaluating the policy, each of the agents must know about the position of the other vehicles in its neighborhood. For each agent to be informed about its own position, a GNSS module is installed on all the drones. The relative angle to the target position, $\psi_D$, can be calculated with knowledge of the target position $\bm{r_D}$ and the current position $\bm{r}$. In order for the other vehicles to be informed about the location of possible obstacles, each vehicle broadcasts its position and heading (required for $\psi_{ji}$) to all other vehicles. XBee radio modules are used for the data transmission, which occurs at a rate of only $3$ Hz. Due to the lack of native support for XBee radio modules, a Raspberry Pi Zero serves as the relay between the Pixhawk and the XBee radio module using the Mavlink protocol. All components relevant for the flight tests are labeled in Fig. \ref{fig:orsula}. The signal path and the inter-drone communication scheme is depicted in Fig. \ref{fig:signal}. Due to space restrictions on the smaller frames, no additional telemetry radios were installed and the communication via the XBee modules provides the link to the ground station. 
		\par The types of commands that can be transmitted from the ground station to the airborne vehicles is limited to basic commands: takeoff, landing, go to a local position (relative to takeoff position), go to a global position, activate hold mode, and emergency shutdown of the motors. The policy is used for all position-based flight modes to avoid collisions. Because of the communication network's mesh architecture, all vehicles can be addressed from one ground-based transmitter.
		\subsection{Flight Test Results}
		\begin{figure}[t!]
			\centering
			\begin{subfigure}[b]{3.0in}
				\centering
				\begin{tikzpicture}

\definecolor{colora}{rgb}{0.12156862745098,0.466666666666667,0.705882352941177}
\definecolor{colora1}{rgb}{1,0.498039215686275,0.0549019607843137}
\definecolor{colora2}{rgb}{0.172549019607843,0.627450980392157,0.172549019607843}

\definecolor{color1}{RGB}{22,55,128}
\definecolor{color2}{RGB}{40,136,255}
\definecolor{color3}{RGB}{7,37,134}
\definecolor{color4}{RGB}{114,136,210}
\definecolor{color5}{RGB}{29,74,112}
\definecolor{color6}{RGB}{101,166,247}
\definecolor{color7}{RGB}{79,153,216}
\definecolor{color8}{RGB}{37,111,179}
\definecolor{color9}{RGB}{13,92,200}
\definecolor{color10}{RGB}{11,47,96}

\begin{axis}[
width=3.0in,
height=1.8in,
tick align=outside,
tick pos=left,
grid style={line width=.25pt, draw=gray!40},
major grid style={line width=.5pt,draw=gray!70},
minor tick num=1,
ylabel={North, m},
xlabel={East, m},
xtick={0,10,20,30,40,50,60,70,80,90,100,110,120},
ytick={-10,0,10,20,30,40,50,60,70,80,90,100,110,120},
xtick style={color=black},
xmin=0,xmax=120,
ymin=0, ymax=60,
ytick style={color=black},
xmajorgrids,
xminorgrids,
ymajorgrids,
yminorgrids,
ticklabel style = {font=\scriptsize}
]
\addplot [semithick, color1,
postaction={decorate, decoration={markings,
mark=at position 0 with{\draw[fill] circle (.3ex);},
	mark=between positions 0.095785 and 1 step 0.095785 with {\arrow{latex'};},
mark=at position 1 with{\draw[fill] circle (.3ex);}
      }}]
table {orsula4.txt};

\addplot [semithick, color1,
postaction={decorate, decoration={markings,
		mark=at position 0 with{\draw[fill] circle (.3ex);},
		mark=between positions 0.080386 and 1 step 0.080386 with {\arrow{latex'};},
		mark=at position 1 with{\draw[fill] circle (.3ex);}
}}]
table {hansbibber4.txt};

\end{axis}

\end{tikzpicture}
				\captionsetup{skip=-0.5mm}
				\caption{Two vehicles, scenario A.}
				\label{fig:flight_tests_2a}
			\end{subfigure}
			\begin{subfigure}[b]{3.0in}
				\centering
				\begin{tikzpicture}

\definecolor{colora}{rgb}{0.12156862745098,0.466666666666667,0.705882352941177}
\definecolor{colora1}{rgb}{1,0.498039215686275,0.0549019607843137}
\definecolor{colora2}{rgb}{0.172549019607843,0.627450980392157,0.172549019607843}

\definecolor{color1}{RGB}{22,55,128}
\definecolor{color2}{RGB}{40,136,255}
\definecolor{color3}{RGB}{7,37,134}
\definecolor{color4}{RGB}{114,136,210}
\definecolor{color5}{RGB}{29,74,112}
\definecolor{color6}{RGB}{101,166,247}
\definecolor{color7}{RGB}{79,153,216}
\definecolor{color8}{RGB}{37,111,179}
\definecolor{color9}{RGB}{13,92,200}
\definecolor{color10}{RGB}{11,47,96}

\begin{axis}[
width=3.0in,
height=1.8in,
tick align=outside,
tick pos=left,
grid style={line width=.25pt, draw=gray!40},
major grid style={line width=.5pt,draw=gray!70},
minor tick num=1,
ylabel={North, m},
xlabel={East, m},
xtick={0,10,20,30,40,50,60,70,80,90,100},
ytick={-10,0,10,20,30,40,50,60,70,80,90,100},
xtick style={color=black},
xmin=0,xmax=70,
ymin=0, ymax=50,
ytick style={color=black},
xmajorgrids,
xminorgrids,
ymajorgrids,
yminorgrids,
ticklabel style = {font=\scriptsize}
]

\addplot [semithick, color1,
postaction={decorate, decoration={markings,
		mark=at position 0 with{\draw[fill] circle (.3ex);},
		mark=between positions 0.079681 and 1 step 0.079681 with {\arrow{latex'};},
		mark=at position 1 with{\draw[fill] circle (.3ex);}
}}]
table {hansbibber2.txt};

\addplot [semithick, color1,
postaction={decorate, decoration={markings,
		mark=at position 0 with{\draw[fill] circle (.3ex);},
		mark=between positions 0.090498 and 1 step 0.090498 with {\arrow{latex'};},	
		mark=at position 1 with{\draw[fill] circle (.3ex);}
}}]
table {cassiopeia2.txt};

\end{axis}

\end{tikzpicture}
				\captionsetup{skip=-0.5mm}
				\caption{Two vehicles, scenario B.}
				\label{fig:flight_tests_2b}
			\end{subfigure}
			\vspace{2mm}
			
			\begin{subfigure}[b]{3.0in}
				\centering
				\begin{tikzpicture}

\definecolor{colora}{rgb}{0.12156862745098,0.466666666666667,0.705882352941177}
\definecolor{colora1}{rgb}{1,0.498039215686275,0.0549019607843137}
\definecolor{colora2}{rgb}{0.172549019607843,0.627450980392157,0.172549019607843}

\definecolor{color1}{RGB}{22,55,128}
\definecolor{color2}{RGB}{40,136,255}
\definecolor{color3}{RGB}{7,37,134}
\definecolor{color4}{RGB}{114,136,210}
\definecolor{color5}{RGB}{29,74,112}
\definecolor{color6}{RGB}{101,166,247}
\definecolor{color7}{RGB}{79,153,216}
\definecolor{color8}{RGB}{37,111,179}
\definecolor{color9}{RGB}{13,92,200}
\definecolor{color10}{RGB}{11,47,96}

\begin{axis}[
width=3.0in,
height=2.8in,
tick align=outside,
tick pos=left,
grid style={line width=.25pt, draw=gray!40},
major grid style={line width=.5pt,draw=gray!70},
minor tick num=1,
ylabel={North, m},
xlabel={East, m},
xtick={0,10,20,30,40,50,60,70,80,90,100},
ytick={-10,0,10,20,30,40,50,60,70,80,90,100},
xtick style={color=black},
xmin=0,xmax=70,
ymin=0, ymax=80,
ytick style={color=black},
xmajorgrids,
xminorgrids,
ymajorgrids,
yminorgrids,
ticklabel style = {font=\scriptsize}
]

\addplot [semithick, color1,
postaction={decorate, decoration={markings,
mark=at position 0 with{\draw[fill] circle (.3ex);},
	mark=between positions 0.085427 and 1 step 0.085427 with {\arrow{latex'};},
mark=at position 1 with{\draw[fill] circle (.3ex);}
      }}]
table {orsula.txt};

\addplot [semithick, color1,
postaction={decorate, decoration={markings,
		mark=at position 0 with{\draw[fill] circle (.3ex);},
		mark=between positions 0.10241 and 1 step 0.10241 with {\arrow{latex'};},
		mark=at position 1 with{\draw[fill] circle (.3ex);}
}}]
table {hansbibber.txt};

\addplot [semithick, color1,
postaction={decorate, decoration={markings,
		mark=at position 0 with{\draw[fill] circle (.3ex);},
		mark=between positions 0.076923 and 0.99 step 0.076923 with {\arrow{latex'};},
		mark=at position 1 with{\draw[fill] circle (.3ex);}
}}]
table {cassiopeia.txt};

\end{axis}

\end{tikzpicture}
				\captionsetup{skip=-0.5mm}
				\caption{Three vehicles.}
				\label{fig:flight_tests_3}
			\end{subfigure}
			\begin{subfigure}[b]{3.0in}
				\centering
				\begin{tikzpicture}

\definecolor{colora}{rgb}{0.12156862745098,0.466666666666667,0.705882352941177}
\definecolor{colora1}{rgb}{1,0.498039215686275,0.0549019607843137}
\definecolor{colora2}{rgb}{0.172549019607843,0.627450980392157,0.172549019607843}

\definecolor{color1}{RGB}{22,55,128}
\definecolor{color2}{RGB}{40,136,255}
\definecolor{color3}{RGB}{7,37,134}
\definecolor{color4}{RGB}{114,136,210}
\definecolor{color5}{RGB}{29,74,112}
\definecolor{color6}{RGB}{101,166,247}
\definecolor{color7}{RGB}{79,153,216}
\definecolor{color8}{RGB}{37,111,179}
\definecolor{color9}{RGB}{13,92,200}
\definecolor{color10}{RGB}{11,47,96}

\begin{axis}[
width=3.0in,
height=2.8in,
tick align=outside,
tick pos=left,
grid style={line width=.25pt, draw=gray!40},
major grid style={line width=.5pt,draw=gray!70},
minor tick num=1,
ylabel={North, m},
xlabel={East, m},
xtick={0,10,20,30,40,50,60,70,80,90,100},
ytick={-10,0,10,20,30,40,50,60,70,80,90,100},
xtick style={color=black},
xmin=0,xmax=80,
ymin=0, ymax=90,
ytick style={color=black},
xmajorgrids,
xminorgrids,
ymajorgrids,
yminorgrids,
ticklabel style = {font=\scriptsize}
]
\addplot [semithick, color1,
postaction={decorate, decoration={markings,
mark=at position 0 with{\draw[fill] circle (.3ex);},
	mark=between positions 0.087413 and 1 step 0.087413 with {\arrow{latex'};},
mark=at position 1 with{\draw[fill] circle (.3ex);}
      }}]
table {orsula3.txt};

\addplot [semithick, color1,
postaction={decorate, decoration={markings,
		mark=at position 0 with{\draw[fill] circle (.3ex);},
		mark=between positions 0.072254 and 1 step 0.072254 with {\arrow{latex'};},
		mark=at position 1 with{\draw[fill] circle (.3ex);}
}}]
table {hansbibber3.txt};

\addplot [semithick, color1,
postaction={decorate, decoration={markings,
		mark=at position 0 with{\draw[fill] circle (.3ex);},
		mark=between positions 0.092251 and 1 step 0.092251 with {\arrow{latex'};},
		mark=at position 1 with{\draw[fill] circle (.3ex);}
}}]
table {cassiopeia3.txt};

\addplot [semithick, color1,
postaction={decorate, decoration={markings,
		mark=at position 0 with{\draw[fill] circle (.3ex);},
		mark=between positions 0.073 and 1 step 0.073 with {\arrow{latex'};},
		mark=at position 1 with{\draw[fill] circle (.3ex);}
}}]
table {gallus3.txt};

\end{axis}

\end{tikzpicture}
				\captionsetup{skip=-0.5mm}
				\caption{Four vehicles.}
				\label{fig:flight_tests_4}
			\end{subfigure}

			\caption{Ground tracks of flight tests for different scenarios with arrows indicating points of equal time for all vehicles.}
			\label{fig:flight_test}
		\end{figure}
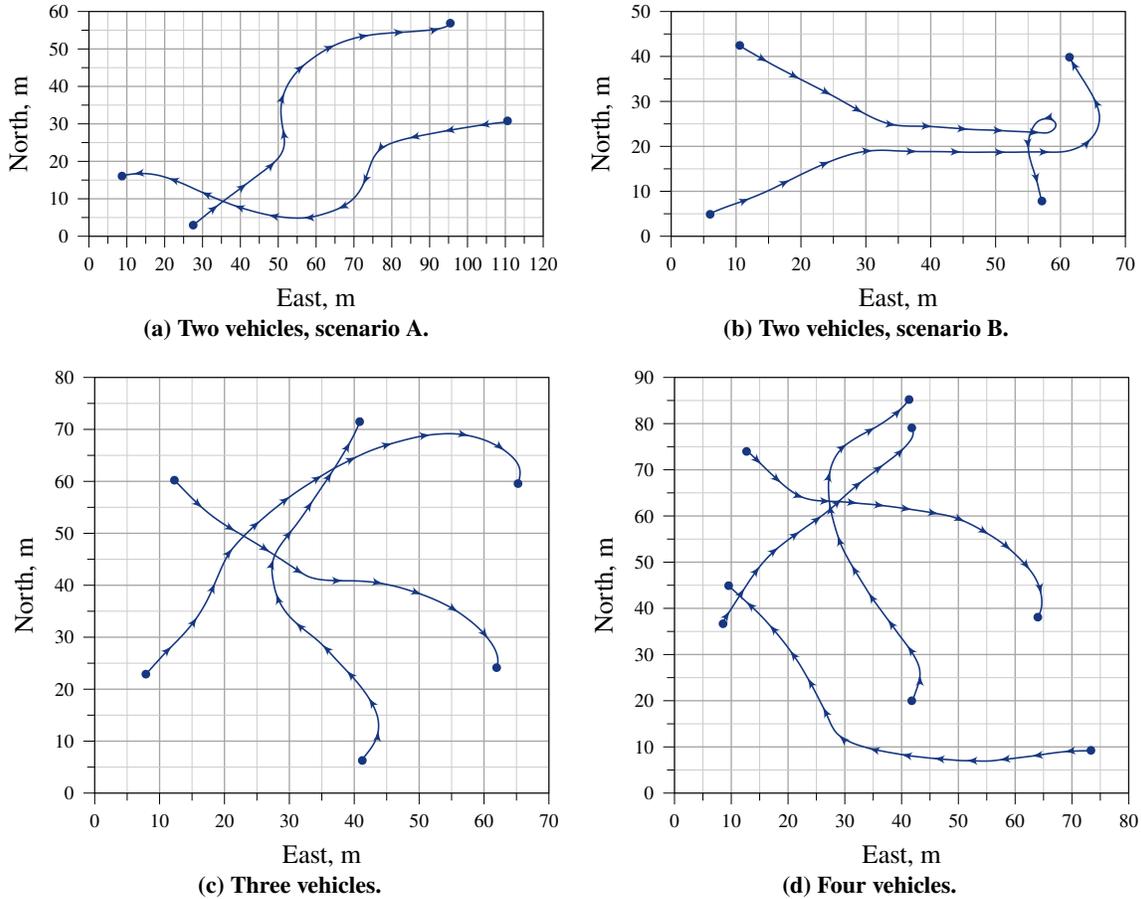
		\par The real-world flight tests compare closely with the results obtained from simulations despite uncertain factors such as weather, sensor imperfection, and communication interruptions. This indicates that the underlying approach is robust with respect to real-world phenomena, despite the fact that the simulation used for the training made various simplifying assumptions. For all scenarios depicted in Fig. \ref{fig:flight_test}, the same policy was used to validate the encoding for a varying number of agents.
		\par Additionally, the algorithm runs on low-cost and low-powered hardware, only taking about 4\% of the Pixhawk's computational resources. The approximately linear behavior in computational complexity as described in section \ref{chap:complexity_analysis} can also be observed when comparing the logging data from different test scenarios.
		\par A general behavioral pattern that can be observed for three or more agents (Figs. \ref{fig:flight_tests_3} and \ref{fig:flight_tests_4}) is the tendency to estimate the future behavior of the other agents, resulting in relatively smooth flight paths. Especially in Fig. \ref{fig:flight_tests_4}, it can be seen that the policy results, for the southernmost agent, in a trajectory that completely avoids the cluttered region around $\bm{r}=\left(30~\text{m},60~\text{m}\right)$. The blocking of the two agents as depicted in Fig. \ref{fig:flight_tests_2b} is a behavior that can occur when the flight direction of the vehicles are approximately equal. For small quantities of vehicles, as in the case provided here, better solutions exist. However, for larger quantities of vehicles, flocking results in more efficient usage of space. Overall, the flight tests show that LSTM-based encoding not only proves functional in a real-world environment with all possible confounding effects, but is also executable on commonly used hardware.
	
	\section{Conclusions}
	\par The proposed approach towards path planning utilizes artificial neural networks and deep reinforcement learning techniques for a continuous observation and action space. To accomplish this procedure, Proximal Policy Optimization (PPO) is used as a training algorithm. The training is expanded to facilitate multi-agent environments though the use of Long Short-Term Memory (LSTMs) as an encoding mechanism. This approach holds significant advantages over the commonly used maps since the information density is higher, the state space remains small, and additional states beyond the position can easily be incorporated into the planning procedure. In fact, the number of states is indefinite. Additionally, LSTMs can naturally deal with an unlimited number of agents, while the number of required floating point operations scales linearly with the number of agents. 
	\par Using a relatively small set of agents for training and a constrained physical space supports the initial success for a converging policy while limiting the computation time required. However, the final policy is still fully capable of dealing with an unlimited number of agents in a physical space of indefinite size. Hence the presented path planning method is particularly suited for environments where high scalability is required. All the characteristic properties of the theoretically developed approach are successfully verified in a real-world multi-agent environment using quadrotor drones.
	
	


	\bibliography{bibfile}

\begin{thebibliography}{41}
\newcommand{\enquote}[1]{``#1''}
\providecommand{\natexlab}[1]{#1}
\providecommand{\url}[1]{\texttt{#1}}
\providecommand{\urlprefix}{URL }
\expandafter\ifx\csname urlstyle\endcsname\relax
  \providecommand{\doi}[1]{\discretionary{}{}{}https://doi.org/#1}\else
  \providecommand{\doi}[1]{\discretionary{}{}{}\urlstyle{rm}\url{https://doi.org/#1}}\fi

\bibitem[{{Katz} et~al.(2019){Katz}, {Le Bihan}, and
  {Kochenderfer}}]{Katz.2019}
{Katz}, S.~M., {Le Bihan}, A., and {Kochenderfer}, M.~J., \enquote{Learning an
  Urban Air Mobility Encounter Model from Expert Preferences,} \emph{2019
  IEEE/AIAA 38th Digital Avionics Systems Conference (DASC)}, 2019, pp. 1--8.
\newblock \doi{10.1109/DASC43569.2019.9081648}.

\bibitem[{{Kuchar} and {Yang}(2000)}]{Kuchar.2000}
{Kuchar}, J.~K., and {Yang}, L.~C., \enquote{A Review of Conflict Detection and
  Resolution Modeling Methods,} \emph{IEEE Transactions on Intelligent
  Transportation Systems}, Vol.~1, No.~4, 2000, pp. 179--189.
\newblock \doi{10.1109/6979.898217}.

\bibitem[{{Beard} et~al.(2006){Beard}, {McLain}, {Nelson}, {Kingston}, and
  {Johanson}}]{Beard.2006}
{Beard}, R.~W., {McLain}, T.~W., {Nelson}, D.~B., {Kingston}, D., and
  {Johanson}, D., \enquote{Decentralized Cooperative Aerial Surveillance Using
  Fixed-Wing Miniature UAVs,} \emph{Proceedings of the IEEE}, Vol.~94, No.~7,
  2006, pp. 1306--1324.
\newblock \doi{10.1109/JPROC.2006.876930}.

\bibitem[{{Kingston} et~al.(2008){Kingston}, {Beard}, and
  {Holt}}]{Kingston.2008}
{Kingston}, D., {Beard}, R.~W., and {Holt}, R.~S., \enquote{Decentralized
  Perimeter Surveillance Using a Team of UAVs,} \emph{IEEE Transactions on
  Robotics}, Vol.~24, No.~6, 2008, pp. 1394--1404.
\newblock \doi{10.1109/TRO.2008.2007935}.

\bibitem[{Lavalle(1998)}]{LaValle.1998}
Lavalle, S.~M., \enquote{Rapidly-Exploring Random Trees: A New Tool for Path
  Planning,} Tech. rep., 1998.

\bibitem[{{Kavraki} et~al.(1996){Kavraki}, {Svestka}, {Latombe}, and
  {Overmars}}]{Kavraki.1996}
{Kavraki}, L.~E., {Svestka}, P., {Latombe}, J.~., and {Overmars}, M.~H.,
  \enquote{Probabilistic Roadmaps for Path Planning in High-Dimensional
  Configuration Spaces,} \emph{IEEE Transactions on Robotics and Automation},
  Vol.~12, No.~4, 1996, pp. 566--580.
\newblock \doi{10.1109/70.508439}.

\bibitem[{Mayne et~al.(2000)Mayne, Rawlings, Rao, and Scokaert}]{Mayne.2000}
Mayne, D., Rawlings, J., Rao, C., and Scokaert, P., \enquote{Constrained Model
  Predictive Control: Stability and Optimality,} \emph{Automatica}, Vol.~36,
  No.~6, 2000, pp. 789 -- 814.
\newblock \doi{10.1016/S0005-1098(99)00214-9}.

\bibitem[{Joos and Fichter(2011)}]{Joos.2011}
Joos, A., and Fichter, W., \enquote{Parallel Implementation of Constrained
  Nonlinear Model Predictive Controller for an FPGA-Based Onboard Flight
  Computer,} \emph{Advances in Aerospace Guidance, Navigation and Control},
  edited by F.~Holzapfel and S.~Theil, Springer Berlin Heidelberg, Berlin,
  Heidelberg, 2011, pp. 273--286.
\newblock \doi{10.1007/978-3-642-19817-5_22}.

\bibitem[{{Ji} et~al.(2017){Ji}, {Khajepour}, {Melek}, and {Huang}}]{Ji.2017}
{Ji}, J., {Khajepour}, A., {Melek}, W.~W., and {Huang}, Y., \enquote{Path
  Planning and Tracking for Vehicle Collision Avoidance Based on Model
  Predictive Control With Multiconstraints,} \emph{IEEE Transactions on
  Vehicular Technology}, Vol.~66, No.~2, 2017, pp. 952--964.
\newblock \doi{10.1109/TVT.2016.2555853}.

\bibitem[{{Bellingham} et~al.(2002){Bellingham}, {Richards}, and
  {How}}]{Bellingham.2002}
{Bellingham}, J., {Richards}, A., and {How}, J.~P., \enquote{Receding Horizon
  Control of Autonomous Aerial Vehicles,} \emph{Proceedings of the 2002
  American Control Conference (IEEE Cat. No.CH37301)}, Vol.~5, 2002, pp.
  3741--3746.
\newblock \doi{10.1109/ACC.2002.1024509}.

\bibitem[{{Richards} and {How}(2002)}]{Richards.2002}
{Richards}, A., and {How}, J.~P., \enquote{Aircraft Trajectory Planning with
  Collision Avoidance Using Mixed Integer Linear Programming,}
  \emph{Proceedings of the 2002 American Control Conference (IEEE Cat.
  No.CH37301)}, Vol.~3, 2002, pp. 1936--1941.
\newblock \doi{10.1109/ACC.2002.1023918}.

\bibitem[{Karaman and Frazzoli(2011)}]{Karaman.2011}
Karaman, S., and Frazzoli, E., \enquote{Sampling-Based Algorithms for Optimal
  Motion Planning,} \emph{The International Journal of Robotics Research},
  Vol.~30, No.~7, 2011, pp. 846--894.
\newblock \doi{10.1177/0278364911406761}.

\bibitem[{Kochenderfer et~al.(2012)Kochenderfer, Holland, and
  Chryssanthacopoulos}]{Kochenderfer.2012}
Kochenderfer, M., Holland, J., and Chryssanthacopoulos, J.,
  \enquote{Next-Generation Airborne Collision Avoidance System,} \emph{Lincoln
  Laboratory Journal}, Vol.~19, No.~1, 2012, pp. 17--33.

\bibitem[{Julian et~al.(2019)Julian, Kochenderfer, and Owen}]{Julian.2018a}
Julian, K.~D., Kochenderfer, M.~J., and Owen, M.~P., \enquote{Deep Neural
  Network Compression for Aircraft Collision Avoidance Systems,} \emph{Journal
  of Guidance, Control, and Dynamics}, Vol.~42, No.~3, 2019, pp. 598--608.
\newblock \doi{10.2514/1.g003724}.

\bibitem[{Watanabe and Johnson(2018)}]{Watanabe.2018}
Watanabe, T., and Johnson, E.~N., \enquote{Trajectory Generation using Deep
  Neural Network,} \emph{2018 {AIAA} Information Systems-{AIAA} Infotech @
  Aerospace}, American Institute of Aeronautics and Astronautics, 2018.
\newblock \doi{10.2514/6.2018-1893}.

\bibitem[{Inoue et~al.(2019)Inoue, Yamashita, and Nishida}]{Inoue.2019}
Inoue, M., Yamashita, T., and Nishida, T., \enquote{Robot Path Planning by LSTM
  Network Under Changing Environment,} \emph{Advances in Computer Communication
  and Computational Sciences}, edited by S.~K. Bhatia, S.~Tiwari, K.~K. Mishra,
  and M.~C. Trivedi, Springer Singapore, Singapore, 2019, pp. 317--329.
\newblock \doi{10.1007/978-981-13-0341-8_29}.

\bibitem[{Notter et~al.(2019)Notter, Zürn, Groß, and Fichter}]{Notter.2019}
Notter, S., Zürn, M., Groß, P., and Fichter, W., \enquote{Reinforced Learning
  to Cross-Country Soar in the Vertical Plane of Motion,} \emph{{AIAA} Scitech
  2019 Forum}, American Institute of Aeronautics and Astronautics, 2019.
\newblock \doi{10.2514/6.2019-1420}.

\bibitem[{Yang and Meng(2000)}]{Yang.2000}
Yang, S.~X., and Meng, M., \enquote{An Efficient Neural Network Approach to
  Dynamic Robot Motion Planning,} \emph{Neural Networks}, Vol.~13, No.~2, 2000,
  pp. 143 -- 148.
\newblock \doi{10.1016/S0893-6080(99)00103-3}.

\bibitem[{Khatib(1986)}]{Khatib.1986}
Khatib, O., \enquote{Real-Time Obstacle Avoidance for Manipulators and Mobile
  Robots,} \emph{The International Journal of Robotics Research}, Vol.~5,
  No.~1, 1986, pp. 90--98.
\newblock \doi{10.1177/027836498600500106}.

\bibitem[{Aamir(2013)}]{Aamir.2013}
Aamir, M., \enquote{On Replacing PID Controller with ANN Wontroller for DC
  Motor Position Control,} \emph{International Journal of Research Studies in
  Computing}, Vol.~2, No.~1, 2013.
\newblock \doi{10.5861/ijrsc.2013.236}.

\bibitem[{Pomerleau(1989)}]{Pomerlau.1989}
Pomerleau, D.~A., \enquote{ALVINN: An Autonomous Land Vehicle in a Neural
  Network,} \emph{Advances in Neural Information Processing Systems}, 1989, pp.
  305--313.

\bibitem[{Richards and Boyle(2010)}]{Richards.2010}
Richards, A., and Boyle, P., \enquote{Combining Planning and Learning for
  Autonomous Vehicle Navigation,} \emph{{AIAA} Guidance, Navigation, and
  Control Conference}, American Institute of Aeronautics and Astronautics,
  2010.
\newblock \doi{10.2514/6.2010-7866}.

\bibitem[{{Qu} et~al.(2009){Qu}, {Yang}, {Willms}, and {Yi}}]{Qu.2009}
{Qu}, H., {Yang}, S.~X., {Willms}, A.~R., and {Yi}, Z., \enquote{Real-Time
  Robot Path Planning Based on a Modified Pulse-Coupled Neural Network Model,}
  \emph{IEEE Transactions on Neural Networks}, Vol.~20, No.~11, 2009, pp.
  1724--1739.
\newblock \doi{10.1109/TNN.2009.2029858}.

\bibitem[{Abbeel et~al.(2010)Abbeel, Coates, and Ng}]{Abbeel.2010}
Abbeel, P., Coates, A., and Ng, A.~Y., \enquote{Autonomous Helicopter
  Aerobatics through Apprenticeship Learning,} \emph{The International Journal
  of Robotics Research}, Vol.~29, No.~13, 2010, pp. 1608--1639.
\newblock \doi{10.1177/0278364910371999}.

\bibitem[{Tran et~al.(2015)Tran, Cross, Motter, Neilan, Qualls, Rothhaar,
  Trujillo, and Allen}]{Tran.2015}
Tran, L.~D., Cross, C.~D., Motter, M.~A., Neilan, J.~H., Qualls, G., Rothhaar,
  P.~M., Trujillo, A., and Allen, B.~D., \enquote{Reinforcement Learning with
  Autonomous Small Unmanned Aerial Vehicles in Cluttered Environments - `After
  all these years among humans, you still haven't learned to smile.',}
  \emph{15th {AIAA} Aviation Technology, Integration, and Operations
  Conference}, American Institute of Aeronautics and Astronautics, 2015.
\newblock \doi{10.2514/6.2015-2899}.

\bibitem[{Julian and Kochenderfer(2017)}]{Julian.2017}
Julian, K.~D., and Kochenderfer, M.~J., \enquote{Neural Network Guidance for
  {UAVs},} \emph{{AIAA} Guidance, Navigation, and Control Conference}, American
  Institute of Aeronautics and Astronautics, 2017.
\newblock \doi{10.2514/6.2017-1743}.

\bibitem[{Matsuura et~al.(2007)Matsuura, Suzuki, Kono, and
  Sakaguchi}]{Matsuura.2007}
Matsuura, A., Suzuki, S., Kono, M., and Sakaguchi, A., \enquote{Lateral
  Guidance Control of UAV Using Feedback Error Learning,} 2007, pp. 174--182.
\newblock \doi{10.2514/6.2007-2727}.

\bibitem[{Sutton and Barto(2018)}]{Sutton.2018}
Sutton, R.~S., and Barto, A.~G., \emph{Reinforcement learning: An
  introduction}, 2\textsuperscript{nd} ed., Adaptive Computation and Machine
  Learning Series, {The MIT Press}, Cambridge Massachusetts, 2018.

\bibitem[{Mnih et~al.(2015)Mnih, Kavukcuoglu, Silver, Rusu, Veness, Bellemare,
  Graves, Riedmiller, Fidjeland, Ostrovski, Petersen, Beattie, Sadik,
  Antonoglou, King, Kumaran, Wierstra, Legg, and Hassabis}]{Mnih.2015}
Mnih, V., Kavukcuoglu, K., Silver, D., Rusu, A., Veness, J., Bellemare, M.,
  Graves, A., Riedmiller, M., Fidjeland, A., Ostrovski, G., Petersen, S.,
  Beattie, C., Sadik, A., Antonoglou, I., King, H., Kumaran, D., Wierstra, D.,
  Legg, S., and Hassabis, D., \enquote{Human-Level Control Through Deep
  Reinforcement Learning,} \emph{Nature}, Vol. 518, No. 7540, 2015, pp.
  529--533.
\newblock \doi{10.1038/nature14236}.

\bibitem[{Julian and Kochenderfer(2019)}]{Julian.2019}
Julian, K.~D., and Kochenderfer, M.~J., \enquote{Distributed Wildfire
  Surveillance with Autonomous Aircraft Using Deep Reinforcement Learning,}
  \emph{Journal of Guidance, Control, and Dynamics}, Vol.~42, No.~8, 2019, pp.
  1768--1778.
\newblock \doi{10.2514/1.g004106}.

\bibitem[{{Ong} and {Kochenderfer}(2015)}]{Ong.2015}
{Ong}, H.~Y., and {Kochenderfer}, M.~J., \enquote{Short-Term Conflict
  Resolution for Unmanned Aircraft Traffic Management,} \emph{2015 IEEE/AIAA
  34th Digital Avionics Systems Conference (DASC)}, 2015, pp. 5A4--1--5A4--13.
\newblock \doi{10.1109/DASC.2015.7311424}.

\bibitem[{Munos and Moore(2002)}]{Munos.2002}
Munos, R., and Moore, A., \enquote{Variable Resolution Discretization in
  Optimal Control,} \emph{Machine Learning}, Vol.~49, No. 2-3, 2002, pp.
  291--323.
\newblock \doi{10.1023/A:1017992615625}.

\bibitem[{Hochreiter and Schmidhuber(1997)}]{Hochreiter.1997}
Hochreiter, S., and Schmidhuber, J., \enquote{Long Short-term Memory,}
  \emph{Neural computation}, Vol.~9, 1997, pp. 1735--80.
\newblock \doi{10.1162/neco.1997.9.8.1735}.

\bibitem[{Brittain and Wei(2021)}]{Brittain2021}
Brittain, M.~W., and Wei, P., \enquote{One to Any: Distributed Conflict
  Resolution with Deep Multi-Agent Reinforcement Learning and Long Short-Term
  Memory,} \emph{AIAA Scitech 2021 Forum}, 2021.
\newblock \doi{10.2514/6.2021-1952}.

\bibitem[{{Kuwata} and {How}(2004)}]{Kuwata.2004}
{Kuwata}, Y., and {How}, J.~P., \enquote{Stable Trajectory Design for Highly
  Constrained Environments Using Receding Horizon Control,} \emph{Proceedings
  of the 2004 American Control Conference}, Vol.~1, 2004, pp. 902--907.
\newblock \doi{10.23919/ACC.2004.1383721}.

\bibitem[{Schulman et~al.(2015)Schulman, Levine, Moritz, Jordan, and
  Abbeel}]{schulman2015trust}
Schulman, J., Levine, S., Moritz, P., Jordan, M.~I., and Abbeel, P.,
  \enquote{Trust Region Policy Optimization,} , 2015.

\bibitem[{Schulman et~al.(2017)Schulman, Wolski, Dhariwal, Radford, and
  Klimov}]{schulman2017proximal}
Schulman, J., Wolski, F., Dhariwal, P., Radford, A., and Klimov, O.,
  \enquote{Proximal Policy Optimization Algorithms,} , 2017.

\bibitem[{Kingma and Ba(2014)}]{Kingma.2014}
Kingma, D.~P., and Ba, J., \enquote{Adam: A Method for Stochastic
  Optimization,} , 2014.

\bibitem[{{Gers} et~al.(1999){Gers}, {Schmidhuber}, and {Cummins}}]{Gers.1999}
{Gers}, F.~A., {Schmidhuber}, J., and {Cummins}, F., \enquote{Learning to
  Forget: Continual Prediction with LSTM,} \emph{1999 Ninth International
  Conference on Artificial Neural Networks ICANN 99. (Conf. Publ. No. 470)},
  Vol.~2, 1999, pp. 850--855.
\newblock \doi{10.1049/cp:19991218}.

\bibitem[{Shannon(1948)}]{Shannon.1948}
Shannon, C., \enquote{A Mathematical Theory of Communication,} \emph{The Bell
  Systems Technical Journal}, Vol.~27, 1948, pp. 379--423,623--656.

\bibitem[{{Meier} et~al.(2015){Meier}, {Honegger}, and
  {Pollefeys}}]{Meier.2015}
{Meier}, L., {Honegger}, D., and {Pollefeys}, M., \enquote{PX4: A Node-Based
  Multithreaded Open Source Robotics Framework for Deeply Embedded Platforms,}
  \emph{2015 IEEE International Conference on Robotics and Automation (ICRA)},
  2015, pp. 6235--6240.
\newblock \doi{10.1109/ICRA.2015.7140074}.

\end{thebibliography}
\end{document}